\documentclass[10pt,twocolumn,letterpaper]{article}

\usepackage{cvpr}              

\usepackage{graphicx}
\usepackage{amsmath}
\usepackage{amssymb}
\usepackage{booktabs}
\usepackage{adjustbox}
\usepackage{multirow}
\usepackage{subcaption}
\usepackage{enumitem}
\usepackage[accsupp]{axessibility}  

\usepackage[pagebackref,breaklinks,colorlinks]{hyperref}

\usepackage[capitalize]{cleveref}
\crefname{section}{Sec.}{Secs.}
\Crefname{section}{Section}{Sections}
\Crefname{table}{Table}{Tables}
\crefname{table}{Tab.}{Tabs.}

\def\cvprPaperID{6260} 
\def\confName{CVPR}
\def\confYear{2023}

\begin{document}

\title{On Calibrating Semantic Segmentation Models: Analyses and An Algorithm}

\author{Dongdong Wang\\
University of Central Florida\\
{\tt\small daniel.wang@knights.ucf.edu}
\and
Boqing Gong\\
Google Research\\
{\tt\small bgong@google.com}
\and
Liqiang Wang\\
University of Central Florida\\
{\tt\small liqiang.wang@ucf.edu}
}
\maketitle

\begin{abstract}

We study the problem of semantic segmentation calibration. Lots of solutions have been proposed to approach model miscalibration of confidence in image classification. However, to date, confidence calibration research on semantic segmentation is still limited. We provide a systematic study on the calibration of semantic segmentation models and propose a simple yet effective approach. First, we find that model capacity, crop size, multi-scale testing, and prediction correctness have impact on calibration. Among them, prediction correctness, especially misprediction, is more important to miscalibration due to over-confidence. Next, we propose a simple, unifying, and effective approach, namely selective scaling, by separating correct/incorrect prediction for scaling and more focusing on misprediction logit smoothing. Then, we study popular existing calibration methods and compare them with selective scaling on semantic segmentation calibration. We conduct extensive experiments with a variety of benchmarks on both in-domain and domain-shift calibration and show that selective scaling consistently outperforms other methods.


\end{abstract}


\let\thefootnote\relax\footnote{CVPR2023 Code/Models: https://github.com/dwang181/selectivecal}

\section{Introduction}
\label{sec:intro}

Deep neural networks (DNNs) have become the ``go-to'' models in various computer vision tasks, such as image classification, object detection, semantic segmentation.
However, recent work found that DNNs are often overconfident when they make mistakes~\cite{guo2017calibration}, misleading downstream applications. 
To calibrate DNNs' confidence in prediction, researchers have developed a rich line of works on image classification using regularized training~\cite{wang2021energy, mukhoti2020calibrating, thulasidasan2019mixup, muller2019does}, post-hoc processing~\cite{guo2017calibration, patel2020multi, kull2019beyond, ma2021meta}, and Bayesian modeling~\cite{lakshminarayanan2017simple, kendall2017uncertainties, izmailov2020subspace, wilson2020bayesian}, to name a few.

\begin{figure}
    \centering
    \includegraphics[width=0.45\textwidth]{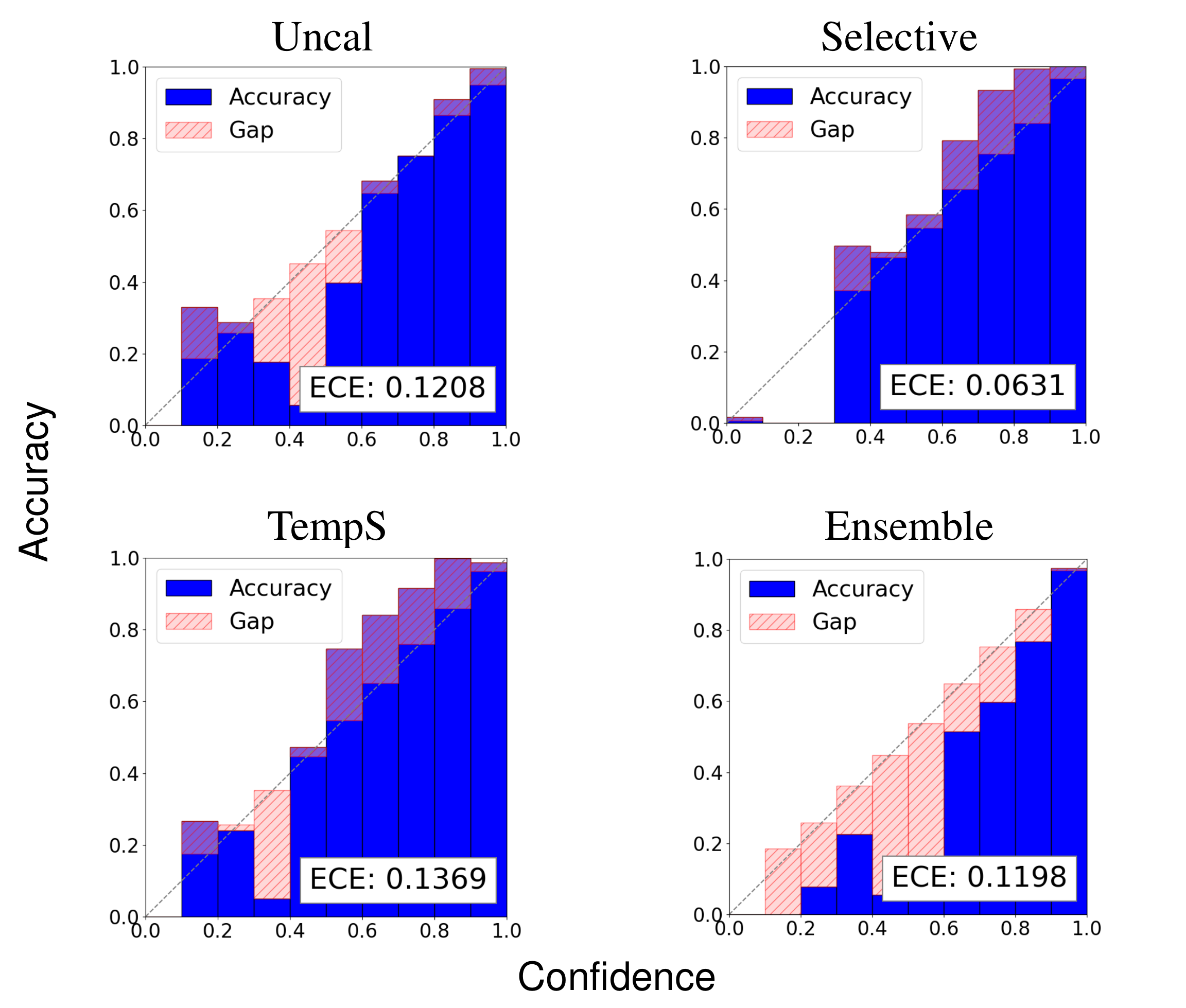}
    \vspace{-0.5\baselineskip}
    \caption{Reliability diagrams \cite{degroot1983comparison} of visualized ECE for Segmenter \cite{strudel2021segmenter}. We randomly select and evaluate fifty validation images from COCO-164K \cite{caesar2018coco} and compare uncalibrated model (Uncal), selective scaling (Selective), temperature scaling (TempS), and ensembling (Ensemble). Smaller gap implies less ECE and better calibration. Ensembling yields comparable accuracy. Despite limited improvement by TempS and Ensemble, Selective exhibits better segmentation model calibration.  }
    \vspace{-1em}
    \label{fig:teaser}
\end{figure}

However, the extensive pursuit of image classification has left it unclear how to calibrate DNNs for other computer vision tasks and how well existing calibration methods generalize to the tasks beyond image classification. In this paper, we conduct a comprehensive study of the calibration of deep semantic segmentation models. 


Semantic segmentation tags a semantic label to every pixel in an image.
Over the past years, we have witnessed increasingly accurate DNN models~\cite{deeplabv3plus2018, zheng2020rethinking, zhang2021knet, fan2021rethinking, strudel2021segmenter, xie2021segformer} for semantic segmentation over various benchmark datasets~\cite{zhou2017scene,caesar2018coco,yu2020bdd100k,pont20172017, van2021multi, isensee2017brain}. 
The progress has benefited many downstream applications, such as medical imaging and diagnostics, autonomous driving, and robotics. While accuracy is essential for the applications, the segmentation models' uncertainties also provide crucial signals, especially for safety-critical applications --- by a segmentation model's {\it uncertainty}, we refer to its confidence in the label it assigns to a pixel. 
For example, an autonomous driving system can use the uncertainty of a semantic segmentation model (e.g., about drivable areas) to make informed decisions. 

Existing work on the semantic segmentation uncertainty focuses on the medical image domain except~\cite{ding2021local}. Deep ensemble~\cite{mehrtash2020confidence, fuchs2021practical, lakshminarayanan2017simple}, MC-dropout~\cite{jungo2019assessing}, and stochastic processes~\cite{monteiro2020stochastic} are introduced from image classification to calibrate a medical image segmentation model's uncertainty. These methods are applied to the training stage and some could reduce a segmentation model's accuracy. We instead focus on the post-hoc calibration methods that do not change the model's prediction, and our study covers various domains. 

Ding et al.'s work~\cite{ding2021local} is the most related to ours. While they proposed and validated a novel calibration method for semantic segmentation, our objective is two-fold. We first investigate the uncertainty in semantic segmentation from multiple perspectives, hoping to reveal the major challenges with the segmentation task's uncertainty. This comprehensive investigation and insights are necessary for future research, given the limited existing work in this area.  We then propose a simple and effective post-hoc algorithm that outperforms and also can apply to almost all existing calibration methods initially designed for image classification, significantly improving their performance in semantic segmentation. 


To fill the gap of current research on semantic segmentation calibration, we conduct a systematic study. We analyze five state-of-the-art models \cite{xie2021segformer, strudel2021segmenter, zhang2021knet, liu2022convnet} with six existing calibration methods \cite{guo2017calibration, kull2019beyond, ma2021meta, ding2021local, mehrtash2020confidence} over seven latest diverse benchmark datasets \cite{zhou2017scene, caesar2018coco, yu2020bdd100k, pont20172017, van2021multi, isensee2017brain, ros2016synthia}. 
First, we investigate different factors to study how they affect miscalibration of semantic segmentation models through experiment. Given the observation, we find that model size, crop size, multi-scaling testing, and misprediction can affect calibration. In particular, segmentation model calibration error is more relevant to misprediction when model size is fixed. Next, we study different existing post-hoc calibrators and introduce our proposed selective scaling algorithm based upon the observation of misprediction. Our selective scaling optimizes ECE by reducing confidence of misprediction, which is reflected by more samples falling into lower confidence region (the far left bin) in Figure \ref{fig:teaser}. Then, we carry out extensive experiments across different models, calibrators, and benchmark datasets. The experiments examine both in-domain calibration and domain-shift calibration to reveal calibration variability across different tasks.


Our contributions can be summarized as follows:

 - We conduct a systematic study on the calibration of semantic segmentation models and provide insights to segmentation model calibration. We find that larger models tend to be less calibrated. When model architecture is fixed, larger crop size and multi-scale testing help calibrate models. Moreover, misprediction is important to miscalibration. 

 - We compare different existing popular calibration methods on state-of-the-art semantic segmentation models, and propose a simple but effective approach, selective scaling, which carries out separate scaling on correct and incorrect predictions. 
 
 - We conduct extensive experiments and justify the effectiveness of selective scaling, as shown in Figure~\ref{fig:teaser}. We also extend experiments from in-domain to domain-shift data and show that selective scaling consistently outperforms.
 
 - We provide useful calibration observations as a comprehensive reference for further research on segmentation model calibration.

\section{Related Work}

\textbf{DNN calibration.} The methods of calibrating DNNs can be categorized into three groups, regularized training, uncertainty estimation, and post-hoc calibration \cite{gawlikowski2021survey}. Regularized training focuses on calibrating DNN over training, such as \cite{wang2021energy, mukhoti2020calibrating, thulasidasan2019mixup, muller2019does}. 
Some methods estimate uncertainty using Bayesian methods \cite{kendall2017uncertainties, wilson2020bayesian, izmailov2020subspace} and deep ensemble \cite{lakshminarayanan2017simple, malinin2019ensemble}. Post-hoc calibration conducts post-processing on the output of DNN to obtain well-calibrated results. While some post-hoc approaches exchange hurting accuracy for better calibration, like histogram binning \cite{guo2017calibration, patel2020multi}, some retain original accuracy of DNNs and only add calibration map to the last layer of DNNs to adjust probability distribution pattern. Several existing binary calibration maps are extended to multiclass calibration. For example, \cite{guo2017calibration} extended temperature scaling and vector scaling aka Logistic scaling\cite{guo2017calibration}, to multiclasss calibration for DNN, and revealed temperature scaling is simple but more effective. To enrich calibration map, Dirichlet scaling\cite{kull2019beyond} was proposed by extending Beta scaling \cite{kull2017beta} to multiclass DNN calibration. Recently, Meta-cal\cite{ma2021meta} integrated bipartite-ranking model with selective classification to improve calibration map for better calibrated model. Despite the success of these existing approaches in image classification calibration, there is no work to discuss their performance on semantic segmentation.  





\textbf{Calibration beyond image classification.} Several extensions based upon existing calibrators have been proposed to solve calibration beyond image classification. For example, \cite{monteiro2020stochastic} modeled predictive uncertainty with spatial correlation matrix in stochastic process and incorporated this information into model training for more calibrated segmentation model. \cite{ding2021local} proposed local temperature scaling that extends temperature scaling to pixels, and thus, better captures spatial heterogeneity of temperature for a better calibrated segmentation model. \cite{mehrtash2020confidence} conducted model ensembling to improve calibration and evaluate the improvement by both conventional and the proposed segment-level uncertainty metrics. \cite{kassapis2021calibrated} addressed semantic segmentation calibration through adversarial training a stochastic network which enables more accurate uncertainty estimation and robust prediction. \cite{fuchs2021practical} proposed a multi-headed Variational U-Net by combining ensemble modeling with variational inference and improves model training with better in-distribution calibration and out-of-distribution detection. \cite{munir2022towards} addressed visual object detector calibration with plug-and-play training-time calibration loss and improved uncertainty optimization. 

\textbf{Domain-shift calibration.} Domain generalization is an important challenge to DNN since model generalization to unseen data could be unpredictable when training and test data distributions differ \cite{torralba2011unbiased}. In terms of model calibration, performance degradation can also be critical \cite{hendrycks2019benchmarking, ovadia2019can}. To alleviate this problem, a variety of approaches have been proposed. \cite{park2020calibrated} corrected covariant shift by importance weighting to achieve better calibrated models. \cite{tomani2021post} proposed a simple yet effective scheme by perturbing image for classification model calibration improvement. \cite{wang2020transferable} proposed TransCal to improve image classification calibration degradation under the scenario of natural distribution shift. \cite{gong2021confidence} leveraged multiple calibration domain to reduce disparity between source and target domain, thus improving model calibration on shift target domain. 
Despite some progress in domain-shift calibration, there is still no domain-shift study on semantic segmentation calibration.

\textbf{Semantic segmentation.} Semantic segmentation is an important visual understanding task. Different from image classification, semantic segmentation focuses on pixel-level classification within an image. Due to this difference, semantic segmentation models are developed in consideration of spatial correlation between pixels. DeepLab family, as an important dilated convolution model, integrates multiscale spatial context into segmentation modeling \cite{deeplabv3plus2018}. SegFormer incorporates hierarchical architectures to encode multiscale features and performs efficient semantic segmentation through ViT structure \cite{xie2021segformer}.
Segmenter is a straightforward ViT extension to semantic segmentation by decoding global context of the encoded embedding features from ViT outputs\cite{strudel2021segmenter}. K-net is a unifying framework for all segmentation tasks \cite{zhang2021knet}. ConvNeXt is proposed as an alternative to ViT and shows accurate segmentation \cite{liu2022convnet}.

\section{Preliminaries}

Similar to image classification, semantic segmentation can be formulated to a multiclass classification problem with a deep neural network. Let $x \in X$ and $y \in Y$ denote an input and its label, respectively. A deep neural network $h(x) = (\hat{y}, \hat{p})$ yields $\hat{y}$ as the predicted label with inference confidence $\hat{p}$. 
We expect a well-calibrated model to provide accurate prediction when its confidence is high. 

\textbf{ECE.} There are several metrics to measure a model's calibration, and one of the most popular and accepted metrics is {\it expected calibration error (ECE)}\cite{naeini2015obtaining}, which reflects the gap between predictive confidence and accuracy. The formal definition with continuous variable is as follows.
\vspace{-0.25em}
\begin{equation}
   ECE = \mathbb{E}_{\hat{p}}\left[|\mathbb{P}(\hat{y} = y | \hat{p} = p) - p|\right], 
   \vspace{-0.25em}
   \label{eq:ECE}
\end{equation}
where $\hat{y}$ is the predicted label, $y$ is the true label, $p$ is expected confidence, and $\mathbb{P}(\hat{y} = y | \hat{p} = p)$ is model predictive accuracy. The expectation is about the discrepancy between accuracy and confidence. A perfectly calibrated model has zero ECE. 

In practical problems, statistical binning is used to quantize continuous variables and estimate eq.~(\ref{eq:ECE}) by equally binning the probability interval,
\vspace{-0.25em}
\begin{equation}
   \widehat{ECE} =\sum_{i=1}^m  \frac{|B_i|}{n}|\mathrm{acc}(B_i) - \mathrm{conf}(B_i)|,
   \vspace{-0.25em}
   \label{eq:ECE_binning}
\end{equation}
where $n$ is the number of samples, $m$ is the number of bins, $B_i$ denotes a set of samples falling into the bin, and acc$(B_i)$ and conf$(B_i)$ are accuracy and confidence averaged over the samples in the bin. $\widehat{ECE}$ can be visualized with the gaps in reliability diagram \cite{degroot1983comparison} as Figure~\ref{fig:teaser}. 

\textbf{ECE in semantic segmentation.} We extend ECE to semantic segmentation by considering each pixel as a sample. Instead of pooling all pixels of different images into a set, we calculate ECE over each image first before taking an average across images,

\vspace{-0.25em}
\begin{equation}
   ECE = \frac{1}{N}\sum_{I=1}^N ECE_{I}, 
   \vspace{-0.25em}
   \label{eq:ECE_img}
\end{equation}
where $I$ is an inference image, and $N$ is the total number of images. Note that we adopt $ECE_{I}$ because an image-wise metric is more popular in segmentation model evaluation. Also, $ECE_{I}$ may significantly vary across images. 







\section{Uncertainty of Segmentation Models}

We analyze the uncertainty of five 
segmentation models, including Segmenter \cite{strudel2021segmenter}, SegFormer \cite{xie2021segformer}, Knet-DeepLab \cite{zhang2021knet}, Knet-SWIN \cite{zhang2021knet}, and ConvNeXt \cite{liu2022convnet}, with the ADE20K \cite{zhou2017scene} dataset, a popular semantic segmentation benchmark with 20,210/2,000 training/validation images. The models are state-of-the-art and released by OpenMMLab~\cite{mmseg2020}. Due to limit of space, we focus on Segmenter~\cite{strudel2021segmenter} and SegFormer~\cite{xie2021segformer} in this section and report other models' results in the supplementary materials. 

\begin{figure*}[h]
    \centering
    \subcaptionbox{\label{sfig:a}}{\includegraphics[width=0.25\textwidth]{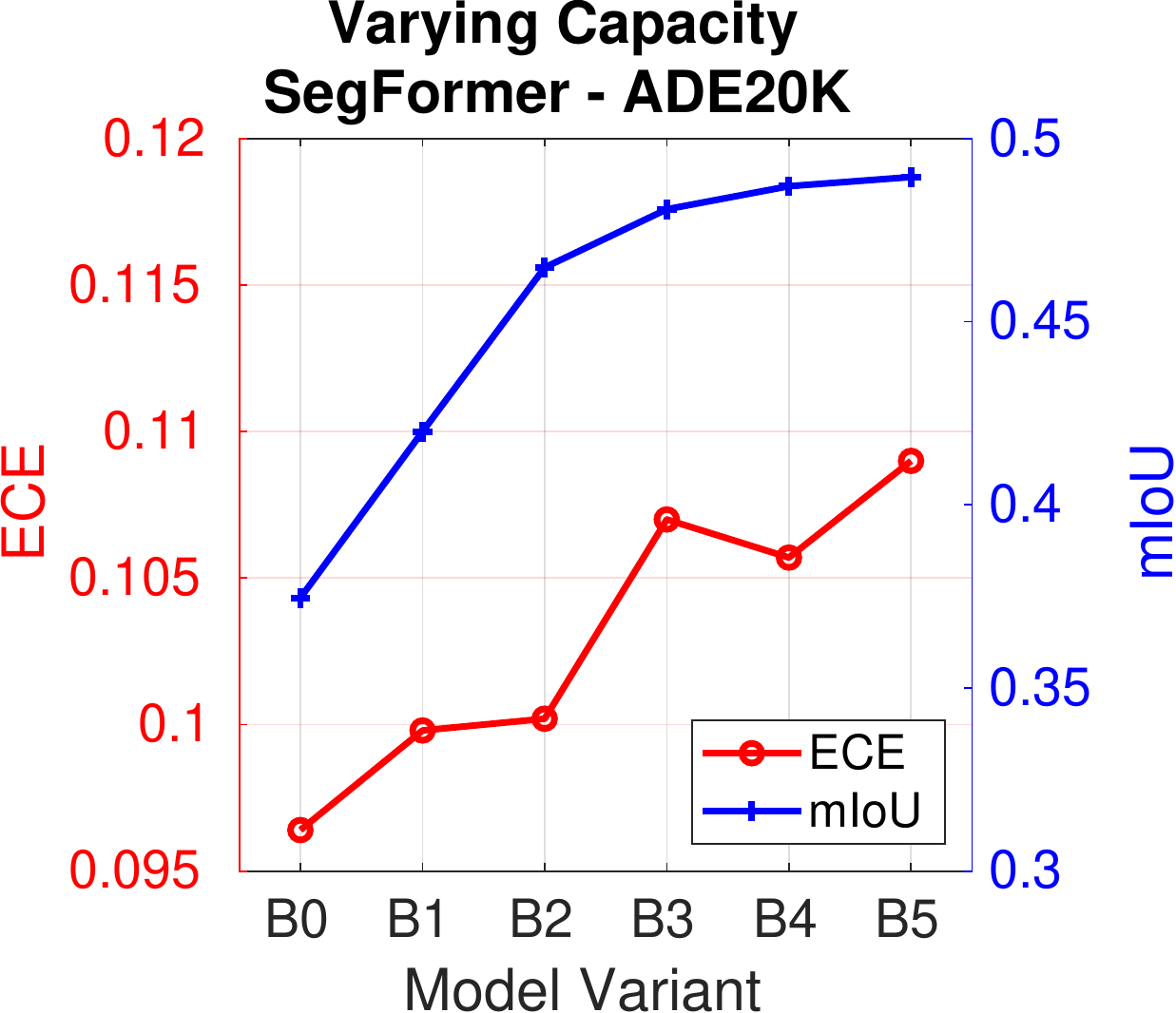}} \hspace{0.25mm}
    \subcaptionbox{\label{sfig:b}}{\includegraphics[width=0.24\textwidth]{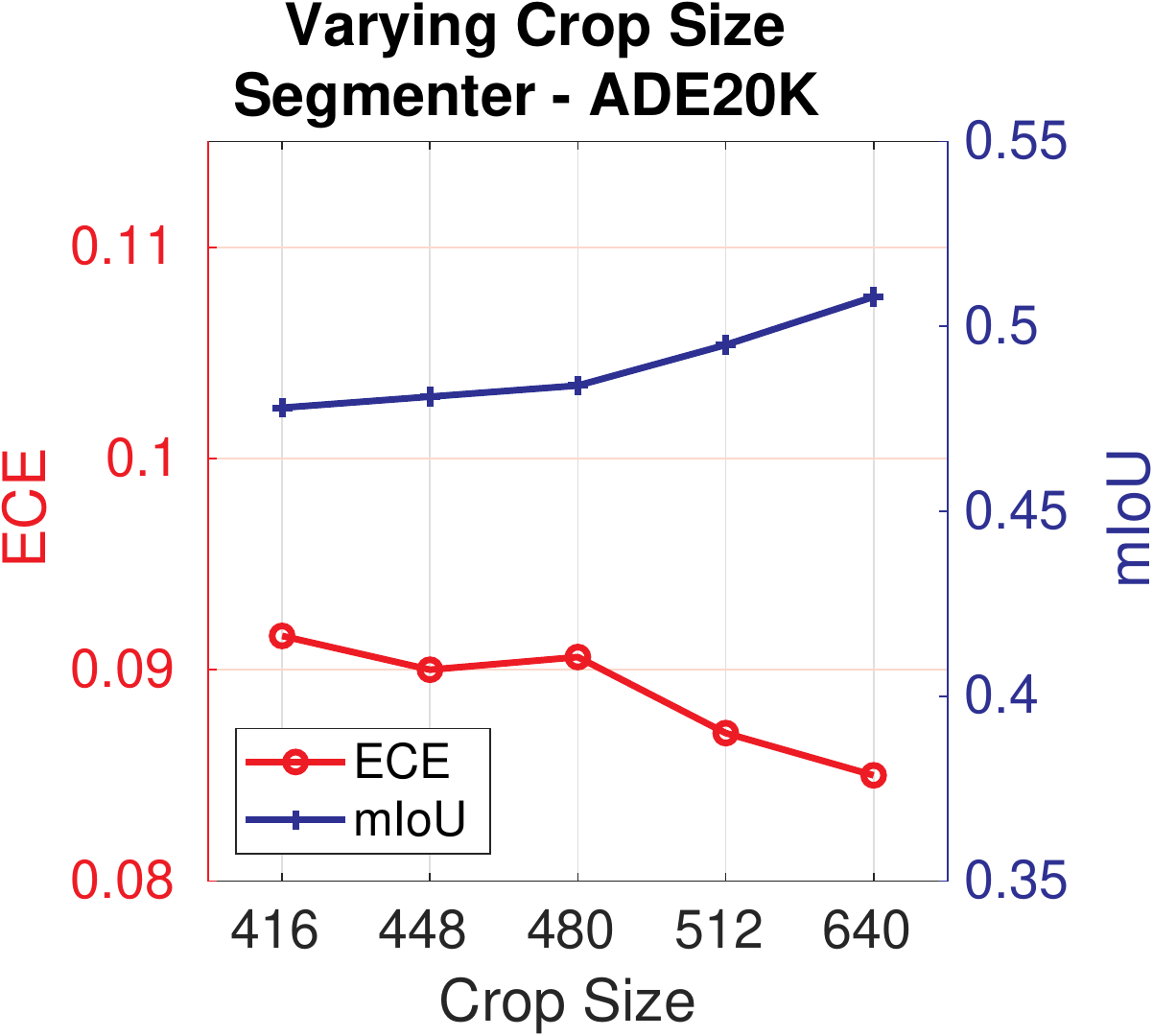}} \hspace{0.25mm}
    \subcaptionbox{\label{sfig:c}}{\includegraphics[width=0.25\textwidth]{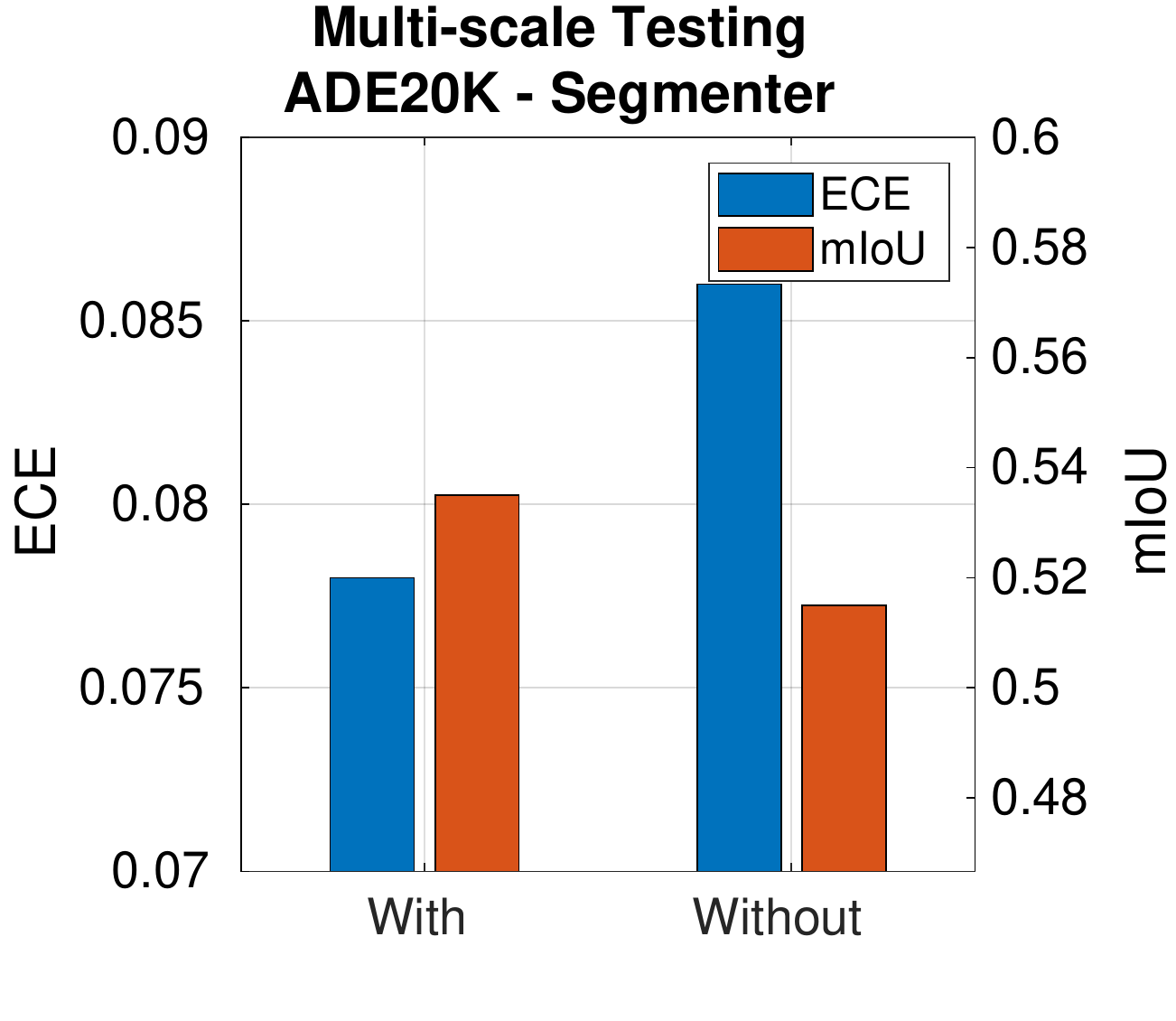}} \hspace{0.25mm}
    \subcaptionbox{\label{sfig:d}}{\includegraphics[width=0.225\textwidth]{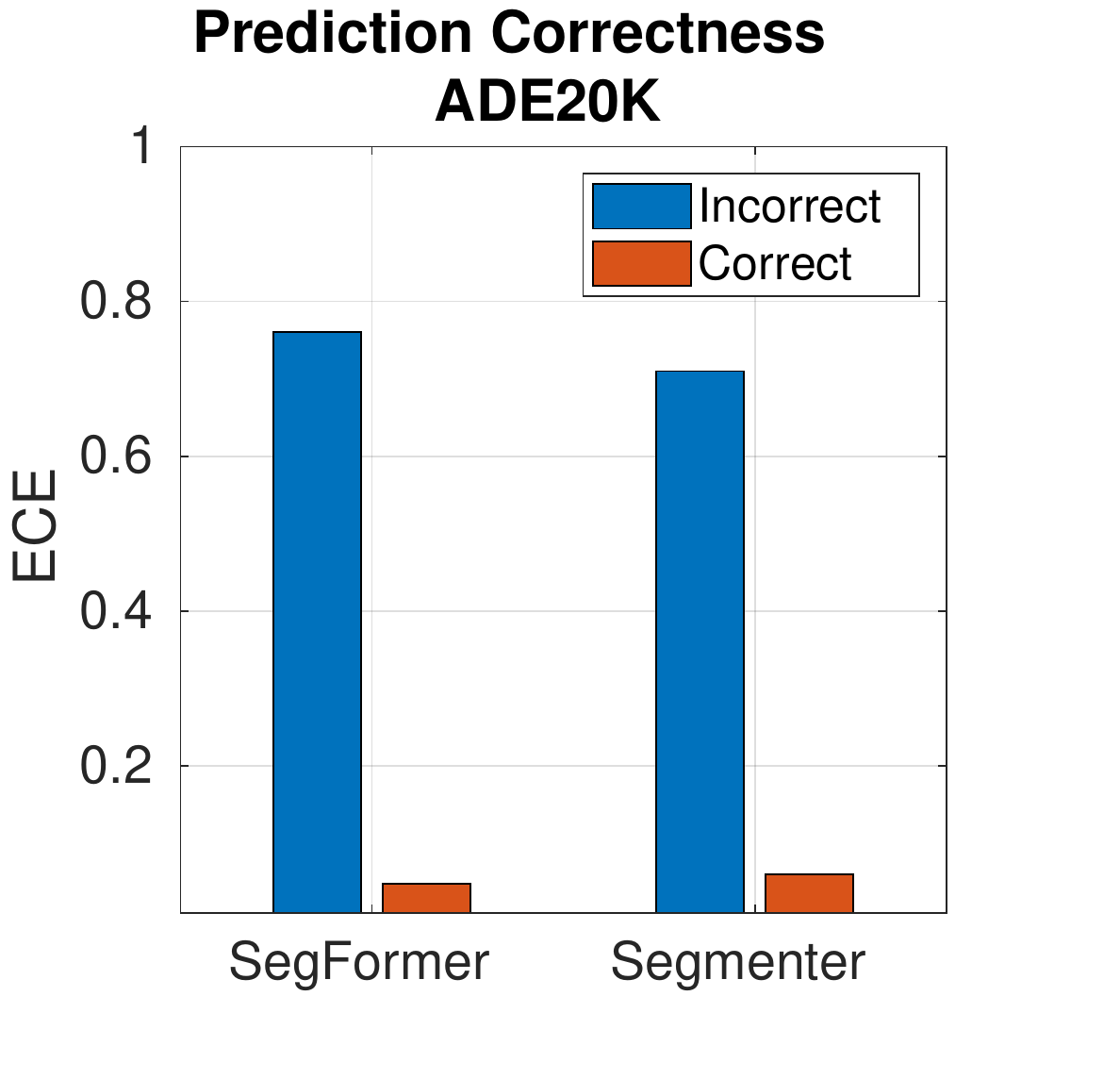}}
     \vspace{-0.75\baselineskip}
    \caption{The effect of model capacity, image crop size, multi-scale testing, and prediction correctness on miscalibration. Prediction correction separately computes the ECE of incorrect/correct predictions. Image-based ECE in eq.(\ref{eq:ECE_img}) is used. }
    \vspace{-1em}
    \label{fig:mis_investigation}
\end{figure*}

\subsection{Larger models tending to be less calibrated} 
Among the models we analyze, we find that the model capacity is a primary factor determining their generalization accuracy. 
Large models often lead to high inference accuracy.
Will it be valid for semantic segmentation models? 
To answer it, we conduct experiments on different model capacity by varying model depth and width. We select five variants of SegFormer\cite{xie2021segformer}, from B0 to B5 with a deeper and wider architecture, for model scaling-up study. From Figure~\ref{sfig:a}, we observe that the deeper and wider models, like B5, exhibit higher ECE (\ie, worse miscalibration) despite better mIoU. It is consistent with the observation for image classification from Figure 2 in \cite{guo2017calibration}, which is caused by the over-confidence during the later stage of NLL optimization \cite{guo2017calibration}. This implies that more accurate prediction leads to worse calibration when it comes to varying model size.

\subsection{Larger crop size tending to be more calibrated}
Image cropping is usually used to prepare input data for semantic segmentation models. Crop size determines model input resolution, \ie, the number of pixels for modeling. Larger crop size indicates larger input image, more pixels fed to model, and wider modeling region. Although larger crop size shows higher predictive accuracy, its effect on model calibration is unknown.  We conduct experiments to explore their interaction. We select Segmenter-L\cite{strudel2021segmenter} as a representative model with five crop sizes of 416, 448, 480, 512, and 640 to show the crop size scaling effect. From Figure~\ref{sfig:b}, we observe that larger crop size yields lower ECE and better calibration. We associate it with modeling field. When crop size increases, larger modeling field provides richer global context information which helps more accurately and confidently locate object boundary and position. This implies that more accurate prediction leads to better calibration in terms of varying crop size.


\subsection{Multi-scaling testing improving calibration}

Multi-scale testing is an important technique to boost semantic segmentation model. It usually resizes the input images into multiple scales, feed these resized images into the model, and average all inferences for final prediction. Predictive accuracy is improved after multi-scale testing because multi-scale features help address scale variation between different objects \cite{zhao2017pyramid}. However, its effectiveness on semantic segmentation calibration is unknown. We conduct experiments with Segmenter-L and cropping size of $640\times640$ of ADE20K to investigate the effect of multi-scale testing. The rescale ratios include 0.5, 0.75, 1.0, 1.25, 1.5, and 1.75. Given Figure~\ref{sfig:c}, with multi-scale testing, ECE is reduced from 0.86 to 0.78 while mIoU increases from 0.515 to 0.535. This suggests that multi-scale testing enables both alleviating miscalibration and improving accuracy. The intuition is that the aggregation of multi-scale features improves the object boundary delineation and predictive confidence in objects. Also, this ensembling helps correct mispredictions and reduce miscalibration over the over-confidence. This implies that more accurate prediction leads to better calibration concerning multi-scaling testing.

\subsection{Mispredictions affecting calibration}
Given eqs.(\ref{eq:ECE}) and (\ref{eq:ECE_binning}), miscalibration results from discrepancy between predictive confidence and accuracy \cite{naeini2015obtaining}. A perfectly calibrated model should yield predicted probability of 1, \ie, 100$\%$ confidence, for correct prediction, and predicted probability of 0, \ie, infinitesimal confidence, for misprediction. However, an optimal solution is hardly obtained in practice, which implies underconfidence in correct prediction and over-confidence in misprediction. How do these discrepancies affect the calibration property of a segmentation model? We conduct an experiment with SegFormer-B5 \cite{xie2021segformer} and Segmenter-L \cite{strudel2021segmenter} by grouping correct/incorrect predictions to separately compute their ECEs. This separation can more clearly reveal how much they contribute to final miscalibration. Figure \ref{sfig:d} shows that the ECEs from incorrect predictions are significantly higher than those from correct predictions across both models. This implies that misprediction more contributes to miscalibration. This observation also justifies that over-confidence is a more critical problem for calibration \cite{guo2017calibration}. 


\section{Calibrating Semantic Segmentation Model}

We first introduce our post-hoc scaling algorithm in this section. Next, we briefly review the popular existing calibration methods, including four image classification post-hoc calibrators, one semantic segmentation calibrator, and one Bayesian modeling calibrator. We select these calibrators because they exhibit state-of-the-art calibration performance in different groups.

\begin{figure*}[h]
    \centering
    \includegraphics[width=0.135\textwidth]{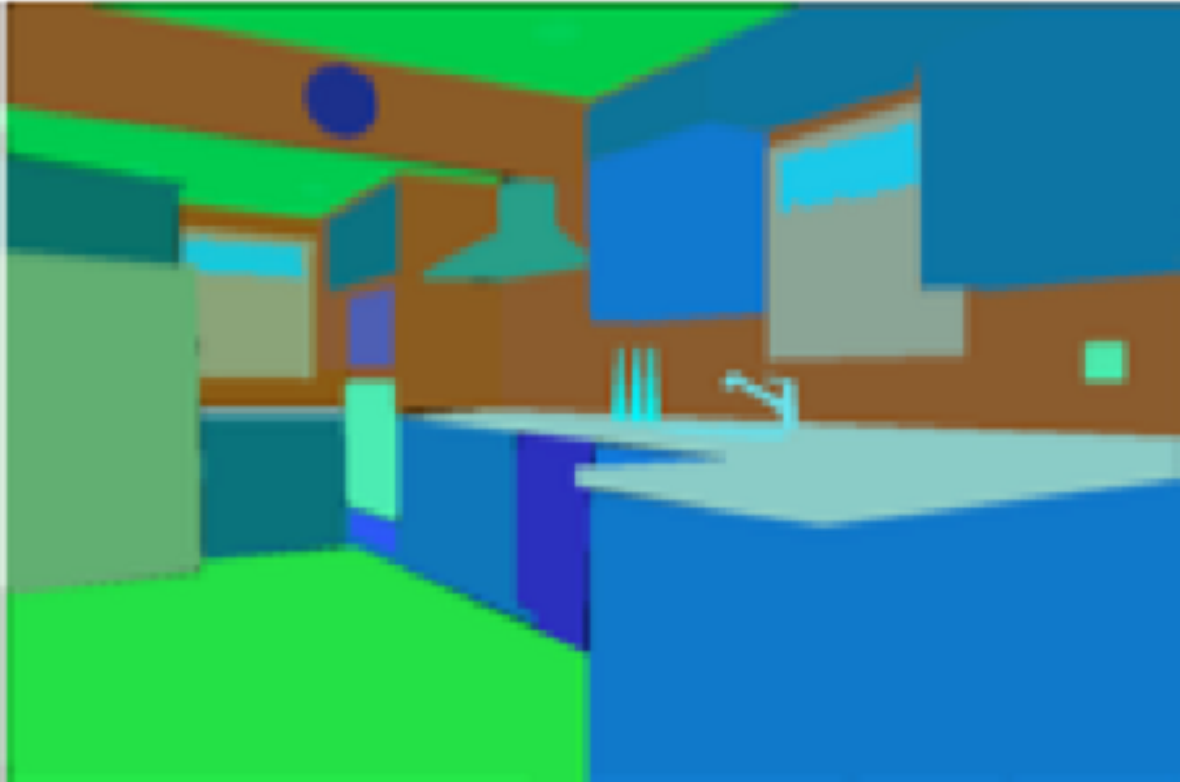} \hspace{1mm}
    \includegraphics[width=0.12\textwidth]{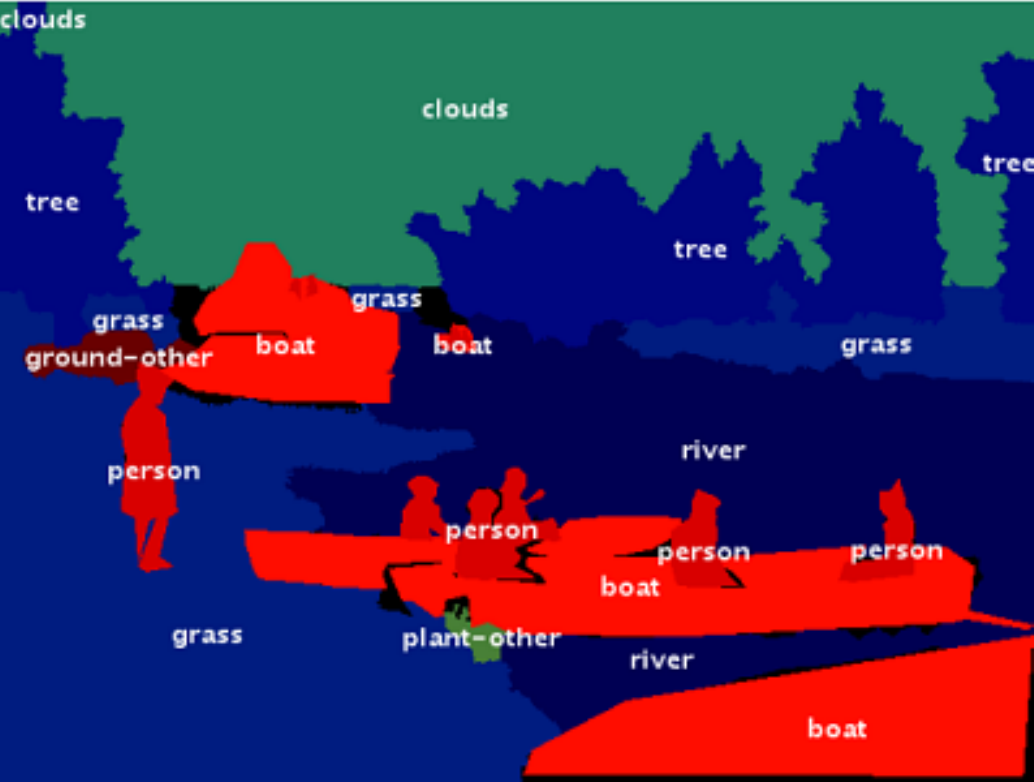} \hspace{1mm}
    \includegraphics[width=0.16\textwidth]{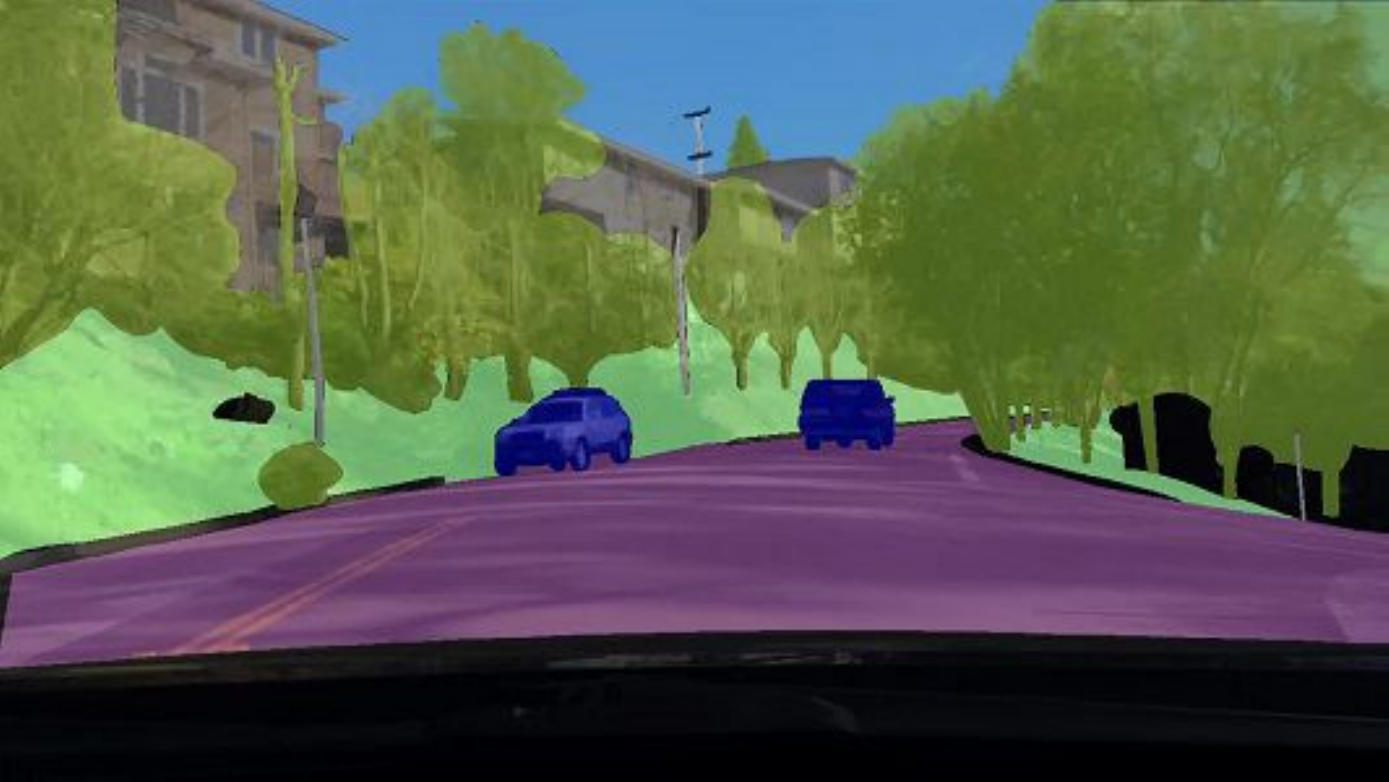} \hspace{1mm}
    \includegraphics[width=0.16\textwidth]{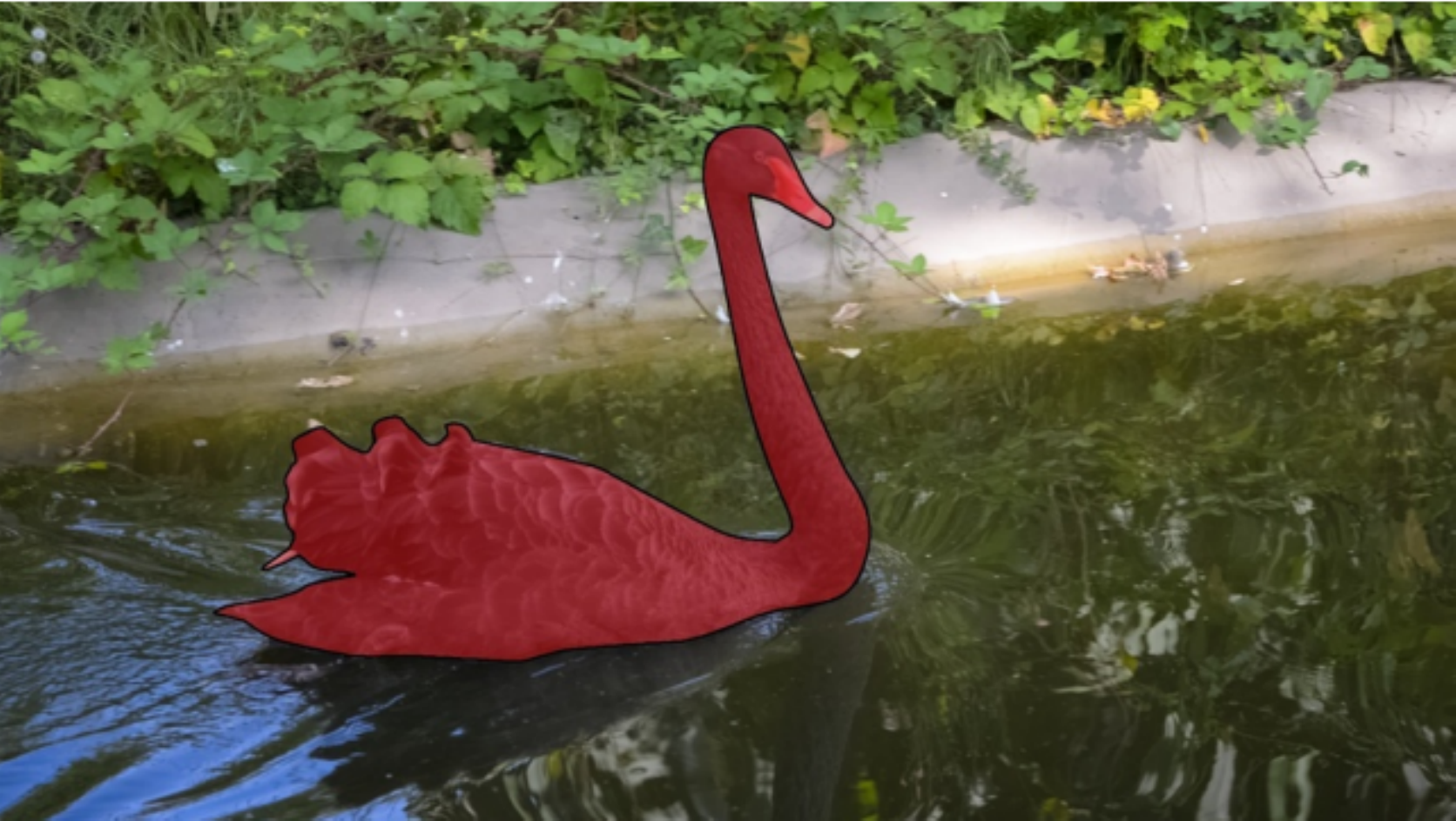} \hspace{1mm}
    \includegraphics[width=0.09\textwidth]{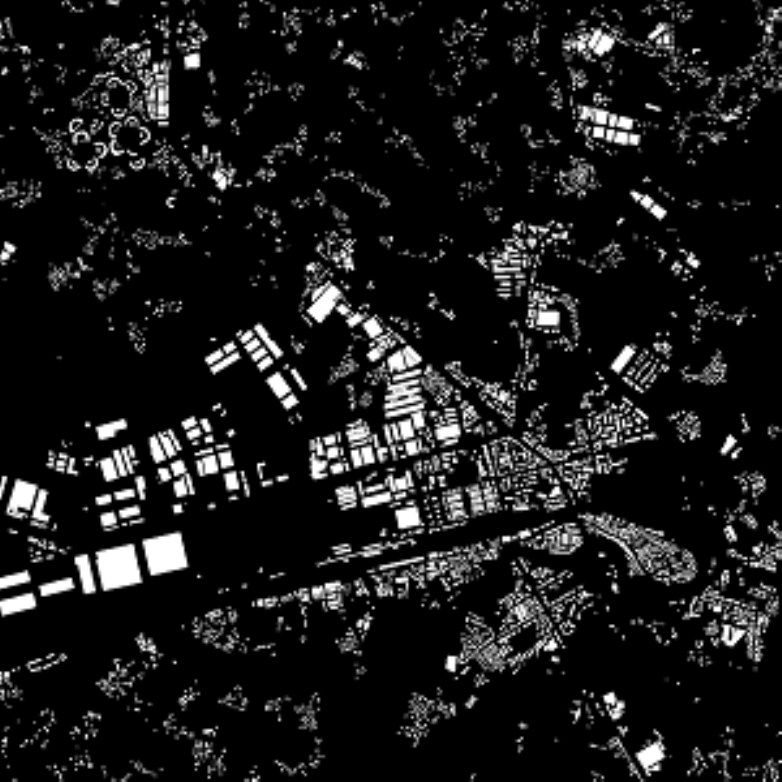} \hspace{1mm}
    \includegraphics[width=0.072\textwidth]{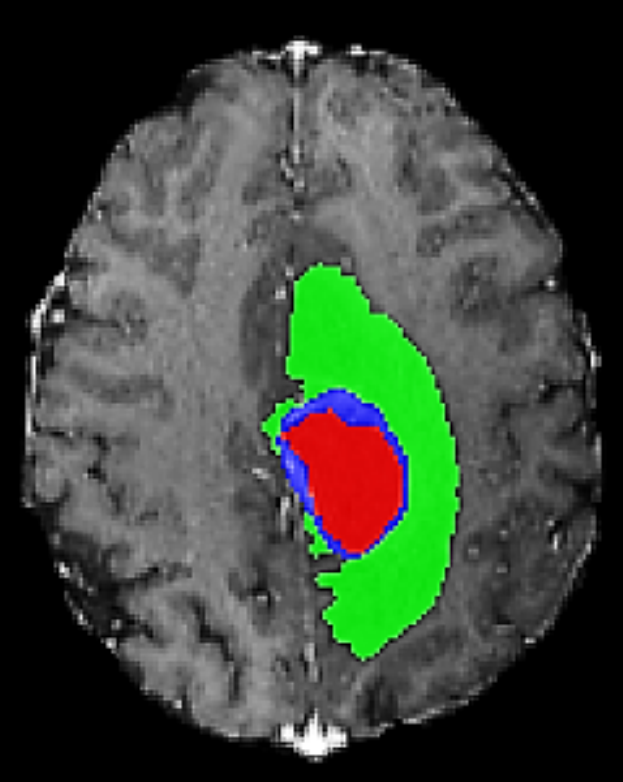} \hspace{1mm}
     \includegraphics[width=0.15\textwidth]{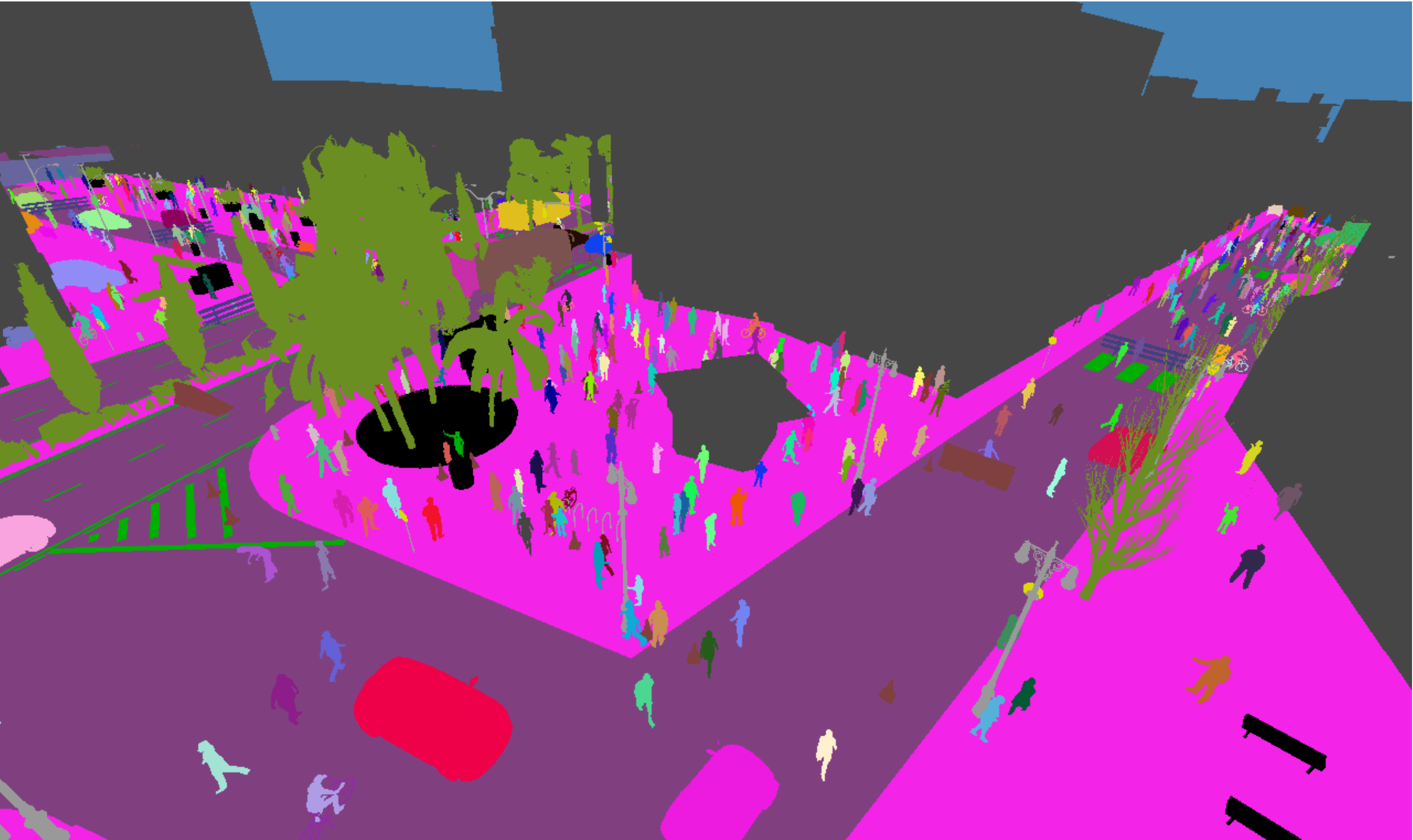}
    \vspace{-0.5\baselineskip}
    \caption{Segmented examples from ADE20K, COCO-164K, BDD100K, DAVIS, SpaceNet-7, BraTs-2017, and SYNTHIA (left to right).}
    \vspace{-0.5em}
    \label{fig:Example}
\end{figure*}

\subsection{Selective Scaling} 

We propose selective scaling given that misprediction is more attributed to miscalibration (Figure \ref{sfig:d}). Selective scaling is an extension of selective classification to model calibration. It introduces a binary classifier as a selector to categorize correct and incorrect predictions for separate scaling. For example, selected mispredictions' logits will be smoothed to alleviate the miscalibration from over-confidence. 

\vspace{-0.25em}
\begin{equation}
\begin{cases}
    \boldsymbol{\hat{p}} = \sigma_{SM}(\boldsymbol{z}/T_1),& \text{if }\hat{y} \neq y\\
    \boldsymbol{\hat{p}} = \sigma_{SM}(\boldsymbol{z}/T_2),              & \text{otherwise}
\end{cases}
\vspace{-0.25em}
\label{eq:Selective}
\end{equation}
where $\boldsymbol{\hat{p}}$ is calibrated probability vector, $\sigma_{SM}$ is softmax activation function, $\boldsymbol{z}$ is the logit vector before activation, $T_1$ is a temperature to smooth logit distribution, and $T_2$ is a temperature to sharpen logit distribution, so $T_1>T_2$. For over-confidence problem, we set $T_2$ to 1. It can be extended to under-confidence problem with $T_2$ less than 1. The selection of $T_1$ relies on the selector accuracy. When its accuracy is higher, $T_1$ is larger. Our selective scaling optimizes ECE by reducing confidence of misprediction, which is reflected by more samples falling into the lower confidence region (the far left bin) in reliability diagram (Figure \ref{fig:teaser}).


Selective scheme was adopted for calibrator design in Meta-cal\cite{ma2021meta}, but different from selective scaling. First, selective scaling does not degrade model accuracy while Meta-cal with miscoverage control does \cite{ma2021meta}. Different from Meta-cal, selective scaling is independent from entropy ranking model to determine classification threshold. Also, selective scaling obtains a selector without heuristic data resampling used in Meta-cal, which reduces hyperparameters. Selective scaling, to the best of our knowledge, is never used for semantic segmentation model calibration. 


\subsection{Existing calibration methods}
\textbf{Temperature Scaling} is a simple but effective approach for multi-classification model calibration \cite{guo2017calibration}. The calibration is carried out with a single temperature parameter to scale logits for overfitting problem resolution. 
\vspace{-0.25em}
\begin{equation}
   \boldsymbol{\hat{p}} = \sigma_{SM}(\boldsymbol{z}/T),
   \vspace{-0.25em}
   \label{eq:TS}
\end{equation}
where $T$ is a scaler of temperature to scale logit vector $\boldsymbol{z}$.

\textbf{Logistic Scaling} is an extension of temperature scaling a.k.a. vector scaling \cite{guo2017calibration}. The scaling model is formulated with linear transformation for more complex calibration map. 
\vspace{-0.25em}
\begin{equation}
   \boldsymbol{\hat{p}} =  \sigma_{SM}(w \odot \boldsymbol{z} + b),
   \vspace{-0.25em}
   \label{eq:LogS}
\end{equation}
where $w$ and $b$ are two vectors to scale the logit vector $\boldsymbol{z}$.

\textbf{Dirichlet Scaling} is the extension of logistic scaling and derived with probability output distribution \cite{kull2019beyond}, which enriches calibration map for better optimization with Dirichlet distribution. We adopt linear paramenterization \cite{kull2019beyond} for Dirichlet scaling, which is formulated as follows.

\vspace{-0.25em}
\begin{equation}
   \boldsymbol{\hat{p}} =  \sigma_{SM}(W \cdot log(\sigma_{SM}(\boldsymbol{z})) + b),
   \vspace{-0.25em}
   \label{eq:LogS}
\end{equation}
where $W$ is a matrix and $b$ is a vector for linear parametrisation of the probability $\sigma_{SM}(\boldsymbol{z})$.

\textbf{Meta-cal} is the scaling strategy derived from bipartite-ranking model and selective classification \cite{ma2021meta}. With the entropy threshold by a ranking model, probability outputs are separately processed. The probability output with the entropy smaller than threshold is scaled by temperature scaling; otherwise, the output is changed to random prediction. Obviously, it severely degrades model accuracy. Although \cite{ma2021meta} proposed coverage accuracy control for image classification, it is impractical to semantic segmentation because mIoU is hard to be tuned with calibrator training set.

\vspace{-0.25em}
\begin{equation}
\begin{cases}
    \boldsymbol{\hat{p}} = \boldsymbol{1}/k,& \text{if } -\hat{p} log(\hat{p})> \gamma , \\
    \boldsymbol{\hat{p}} = \sigma_{SM}(\boldsymbol{z}/T),              & \text{otherwise}, 
\end{cases}
\vspace{-0.25em}
\label{eq:meta}
\end{equation}
where k is the number of classes, $\gamma$ is the entropy threshold computed with a ranking model for calibration decision.

\textbf{Local Temperature Scaling} (LTS) is the extension of temperature scaling to semantic segmentation \cite{ding2021local}. It provides pixel-level heterogeneous scaling temperature by combing input images with output logit through CNN. Although it considers spatial heterogeneity over an image for scaling, the calibrator training relies on strong correlation between image features and temperature heterogeneity and more training data for feature extraction \cite{ding2021local}.
\vspace{-0.25em}
\begin{equation}
   \boldsymbol{\hat{p}} = \sigma_{SM}(\boldsymbol{z}/(\boldsymbol{T}(\boldsymbol{z}, I)),
   \vspace{-0.25em}
   \label{eq:LTS}
\end{equation}
where $\boldsymbol{T}(\boldsymbol{z}, I)$ is a pixel-wise temperature map by a convolution with segmented logit vector $\boldsymbol{z}$ and image $I$.

\textbf{Ensembling}  is proposed to solve medical image segmentation  uncertainty prediction \cite{mehrtash2020confidence}. It carries out calibration improvement with simple average of ensemble model, a simplified version of Bayesian inference \cite{wilson2020bayesian}. Also, deep ensemble is a strong baseline for image classification calibration \cite{lakshminarayanan2017simple, wilson2020bayesian}. We compare its performance with other post-hoc approaches. 

\vspace{-0.25em}
\begin{equation}
   \boldsymbol{\hat{p}} = \frac{1}{N} \Sigma_{n=1}^N \boldsymbol{\hat{p}}^n,
   \vspace{-0.25em}
   \label{eq:ensembling}
\end{equation}
where $N$ is the number of ensemble members.



\section{Experiments}

\textbf{Models.} 
We consider five recent state-of-the-art segmentation models which cover CNN and ViT architectures.

\begin{enumerate}[nolistsep]
\item \textbf{SegFormer}\cite{xie2021segformer} is a ViT-based encoder-decoder model based upon lightweight multilayer perception (MLP) decoders with pyramid architecture.
\item \textbf{Segmenter}\cite{strudel2021segmenter} is an encoder-decoder model based exclusively on Transformer.
\item \textbf{Knet} \cite{zhang2021knet} is a unified segmentation decoder module which includes two important variants, Knet-DeepLab and Knet-SWIN. Knet-DeepLab consists of ResNet-50 \cite{he2016deep} backbone and DeepLab-V3 \cite{deeplabv3plus2018} decoder. Knet-SWIN is contructed with SWIN  \cite{liu2021swin} backbone and UperNet \cite{xiao2018unified} decoder.
\item \textbf{ConvNeXt} \cite{liu2022convnet} is a CNN backbone, but shaped to ViT-like architecture for better scaling. We select ConvNeXt backbone with UperNet \cite{xiao2018unified} decoder.

\end{enumerate}

For ADE20K, all models except Knet-DeepLab are trained with 640$\times$640 crop size. For BDD100K and SYNTHIA, the crop size is 512$\times$1024. For other benchmarks, the crop size is 512$\times$512. The batch size is set to 8.

\textbf{Dataset.} We examine existing calibration methods across six important benchmarks from various applications. Figure~\ref{fig:Example} illustrates the example images from benchmarks. 

\begin{enumerate}[nolistsep]

\item \textbf{Scene and stuff segmentation.} We use ADE20K \cite{zhou2017scene} and COCO-164K \cite{caesar2018coco} as large-scale segmentation benchmark. ADE20K contains 150 object and stuff classes with 20,210/2,000 images in the training/validation set. COCO-164K, $\thicksim$164K images for 91 stuff and 80 thing classes, includes 118K/5K images in the training/validation set.

\item \textbf{Autonomous driving.} We choose BDD-100K \cite{yu2020bdd100k}, a latest benchmark for urban driving scene segmentation. BDD-100K contains 7K/1K 1280$\times$720 images in 19 classes for training/validation set. We select the validation set of CityScapes \cite{cordts2016cityscapes} as test set from target domain for domain-shift calibration assessment. CityScapes has 2975/500 1024$\times$2048 images for training/validation set. BDD-100K and CityScapes share the same label space.

\item \textbf{Video segmentation.} We adopt DAVIS2016 \cite{perazzi2016benchmark} to assess calibration on temporal domain shift problems. DAVIS2016 contains 50 object and 1 background classes. We select 480p resolution video sequences to prepare temporal domain shift dataset. The total of the frames in each sequence ranges from 24 to 103. Each frame is an image segmented with a single object and background.

\item \textbf{Remote sensing.} We select SPACENET-7 (SN-7) \cite{van2021multi}, a binary small object identification benchmark, to examine spatial and temporal domain shift calibration. SN-7 consists of 1,408 1024$\times$1024 satellite images collected from monthly urban development across 101 cities. It contains 11,080,000 buildings over 41,000 km$^2$ observation area. Due to the development, images vary temporally; due to planning difference, the objects like the building styles differ by locations.

\item \textbf{Medical imaging.} We select BraTS2017 \cite{isensee2017brain}, a popular brain tumor segmentation dataset. BraTS2017 contains multimodal magnetic resonance imaging (MRI) scans from 1,210 patients, which consists of 3D voxels with 155 layers from top to bottom. Each layer is equivalent to a 2D semantic segmentation image with 3 tumor feature classes. 

\item \textbf{Simulation domain transfer.} SYNTHIA \cite{ros2016synthia} is a popular unban street synthetic image benchmark. We select its subset of RANDCITYSCAPES, which contains 9400 1280$\times$760 images with 22 classes, as source domain for model training. We select CityScapes \cite{cordts2016cityscapes} as target domain for domain-shift calibration assessment.

\end{enumerate}



\textbf{In-domain calibration.} 
We select the validation set of each benchmark for in-domain calibration study. For ADE20K, COCO-164K, and BDD100K, we select 100/100/1800 images, 250/250/4500 images, and 50/50/900 images for calibrator training/validation/testing. 

\textbf{Domain-shift calibration.} We select the validation sets of COCO-164K and CityScapes as testing sets of ADE20K and BDD100K. For taxonomy incompatibility over datasets, like ADE20K-COCO164K, we follow the convention in \cite{lambert2020mseg} to merge and split label classes. For DAVIS, we merge all training and validation sequences and resplit them in half by frame. The first half of frames are used for training with 1731 images and the second half are split for validation/testing with 100/1624 images. For SN-7, all training data are split into 1133/50/225 images for training/validation/testing set by location, and 698/50/660 images for training/validation/testing set by time. For BraTS, we select 800/50 images of 75$^{th}$ layer scan for training/validation and 360 images of 100$^{th}$ layer scan for testing. For SYNTHIA, we split training set into 9000/400 images for training/validation, and CityScapes' validation set for testing.

\textbf{Calibration methods.} We select aforementioned six existing calibration methods to compare selective scaling. For selective scaling, selective model is simplified with three-layer MLP due to training data restriction. The inputs are pixel-wise predictive probabilities and the labels are binary which indicate correct/incorrect prediction. Moreover, Meta-cal with miscoverage control yields severe model degradation. We implement our extension by replacing random prediction with large temperature scaling to retain accuracy. For ensembling, we achieve a three-member ensemble with comparable accuracy for fair comparison.

\begin{table*}[h]
\footnotesize
\centering
\setlength\tabcolsep{5pt}
\caption{Segmentation accuracy (mIoU) and calibration error (ECE) on different benchmarks. TempS, LogS, DirS, LTS, Ens., and Selective denote temperature, logistic, Dirichlet, local temperature, ensembling, and selective scaling, respectively. Ensembling is carried out by three models with reduced size for comparable mIoU. Meta-Cal$^*$ is implemented by our extension with large temperature. }
\label{tab:SoTA_Models}
\vspace{-0.1in}
\begin{tabular}{c c c c c c c c c c c c }
 \hline
 Dataset & Model & mIoU & Uncal& TempS & LogS & DirS & Meta-Cal$^{*}$ & LTS & Ens. & Selective \\
 \hline
ADE20K & SegFormer-B5 \cite{xie2021segformer} & 49.13 & 0.111 & 0.109 & 0.110 & 0.110 & 0.103  & 0.105 & 0.109 & \textbf{0.086}\\ 
ADE20K & Segmenter-L \cite{strudel2021segmenter} &  51.65 & 0.087 & 0.086  & 0.086 & 0.087 & 0.081 & 0.094 & 0.086 & \textbf{0.069}\\ 
ADE20K & Knet-DeepLab \cite{zhang2021knet}&  45.06 & 0.111 & 0.105 & 0.107 & 0.106 & 0.102 &   0.118 & 0.110 & \textbf{0.095}\\
ADE20K & Knet-SWIN-L \cite{zhang2021knet} & 52.46 & 0.098 & 0.094 & 0.093 & 0.097 & 0.089 &  0.134 & 0.096 &\textbf{0.078} \\
ADE20K & ConvNeXt-L \cite{liu2022convnet} & 53.16 &  0.097 & 0.092 & 0.094 & 0.091 & 0.088 &  0.133 & 0.094 &\textbf{0.082} \\
 \hline
COCO-164K & SegFormer-B5 \cite{xie2021segformer} & 45.78 & 0.151 & 0.149 & 0.141 & 0.151 & 0.132 &0.151 & 0.149 &  \textbf{0.113} \\
COCO-164K & Segmenter-L \cite{strudel2021segmenter}&  47.09 & 0.152 & 0.149 & 0.149 & 0.151 & 0.130 &0.155 & 0.150 & \textbf{0.109}\\
COCO-164K & Knet-DeepLab \cite{zhang2021knet}&  37.24 & 0.170 & 0.170 & 0.168 & 0.171 & 0.149  &0.172 & 0.169 & \textbf{0.093}\\
COCO-164K & Knet-SWIN-L \cite{zhang2021knet} & 46.49 & 0.161 & 0.159 & 0.161 & 0.160 & 0.142 &0.162 & 0.160 & \textbf{0.102}\\
COCO-164K & ConvNeXt-L \cite{liu2022convnet} & 46.48 & 0.160 & 0.157 & 0.158 & 0.159 & 0.141 & 0.162 & 0.159 & \textbf{0.108}\\
 \hline
BDD100K & SegFormer-B5 \cite{xie2021segformer} & 65.08 & 0.064  & 0.055  & 0.054 & 0.053 & 0.049 & 0.069  & 0.059  & \textbf{0.040} \\
BDD100K & Segmenter-L \cite{strudel2021segmenter}&  61.33 & 0.055 & 0.045 & 0.043 & 0.042 & 0.037 & 0.071 &  0.052 & \textbf{0.031} \\
BDD100K & Knet-DeepLab \cite{zhang2021knet}&  62.89 & 0.060 &  0.049 & 0.047 &  0.048  & 0.041 & 0.063 & 0.057 & \textbf{0.035} \\
BDD100K & Knet-SWIN-L \cite{zhang2021knet} & 67.59 & 0.065  & 0.055  & 0.054 & 0.054 & 0.049 & 0.067 & 0.063 & \textbf{0.040} \\
BDD100K & ConvNeXt-L \cite{liu2022convnet} & 67.26 & 0.064 & 0.054 & 0.053 & 0.056 & 0.049 & 0.065 & 0.063 & \textbf{0.038} \\

 \hline
\end{tabular}
\end{table*}

\subsection{Results}
\subsubsection{In-domain calibration evaluation}
We present ECEs for five models and compare selective scaling with six existing calibrators on three-run average. Table \ref{tab:SoTA_Models} shows that majority of calibrators yield very limited improvement in model calibration. However, surprisingly, selective scaling consistently outperforms these scaling approaches. This calibration gain is attributed to more scaling on mispredictions. Moreover, this improvement is more significant for scene and object benchmarks such as ADE20K and COCO-164K than street view datasets like BDD-100K.

Furthermore, we find that BDD100K yields more calibrated models than other benchmarks. We associate it with larger crop size ($512\times1024$) and higher model accuracy. We also find that ensembling exhibits weaker calibration. We connect it with model accuracy since we restrain the ensemble accuracy for fair comparison, which limits its calibration performance. Moreover, across models, Segmenter exhibits better calibration. Since it is a Transformer-exclusive model, we link it to different spatial inductive bias which is speculated in \cite{minderer2021revisiting}.



\vspace{-1em}

\subsubsection{Domain-shift calibration evaluation}

We present ECEs for five models on seven sets of domain-shift experiments and compare selective scaling with six existing calibrators on three-run average. Table \ref{tab:domain_shift} shows that selective scaling still consistently outperforms by a large margin. It implies that selective scaling can be generalized to domain-shift problems. In particular, Meta-cal exhibits performance variation. We speculate that the hard threshold from entropy calculation limits separating shift data. In contrast, selective scaling is more consistent because binary classifier has better generalization to separating shift data.

Across benchmarks, we find that label space shift causes more significant miscalibration. For example, ADE20K to COCO-164K and SYNTHIA to CityScapes show worse ECE than other domain-shift experiments. We link it to poorer domain adaptation and model generalization degradation. Also, we observe that small object segmentation like SN-7, as a challenged task, can cause worse miscalibration. Except for SN-7, more accurate models are more calibrated by benchmarks. For SN-7, background pixels engage in training but are excluded from calibration calculation, which causes larger ECE. Within these models, we find that Segmenter is still more calibrated.




\begin{table*}[h]
\footnotesize
\centering
\setlength\tabcolsep{4pt}
\caption{Segmentation model accuracy (mIoU) and calibration error (ECE) on different benchmarks for domain-shift calibration assessment. InD and SD denotes in-domain and domain-shift test data. Meta-Cal$^*$ is our extension with large temperature scaling.  }
\label{tab:domain_shift}
\vspace{-0.1in}
\begin{tabular}{c c c c c c c c c c c c c}
 \hline
 InD & SD & Model & mIoU & Uncal& TempS & LogS & DirS & Meta-Cal$^{*}$ & LTS & Ens. & Selective \\
 \hline
ADE20K & COCO-164K & SegFormer-B5 \cite{xie2021segformer}& 8.20 & 0.467 & 0.466 & 0.468 & 0.469 & 0.415 & 0.420 & 0.465 & \textbf{0.331} \\
ADE20K & COCO-164K & Segmenter-L \cite{strudel2021segmenter}& 9.60 & 0.395 & 0.394 & 0.395 & 0.394 & 0.322 & 0.331 & 0.393 &\textbf{0.259}  \\
ADE20K & COCO-164K & Knet-DeepLab \cite{zhang2021knet} & 7.02 & 0.447 & 0.445 & 0.446 & 0.446 & 0.403 & 0.439 & 0.444 &\textbf{0.351} \\
ADE20K & COCO-164K & Knet-SWIN-L \cite{zhang2021knet} & 9.06  & 0.464 & 0.462 & 0.463 & 0.463& 0.379 &  0.465 & 0.462 &\textbf{0.327}\\
ADE20K & COCO-164K & ConvNeXt-L \cite{liu2022convnet} & 8.99 & 0.461 & 0.458 & 0.459 & 0.460 & 0.377 & 0.460 & 0.459 &\textbf{0.320}\\
 \hline
BDD100K & CityScapes & SegFormer-B5 \cite{xie2021segformer}& 63.05 & 0.087 & 0.083  & 0.082  & 0.084 & 0.076  & 0.085 & 0.087 & \textbf{0.069} \\
BDD100K & CityScapes &Segmenter-L \cite{strudel2021segmenter}& 60.26  & 0.062  & 0.058 & 0.059 &  0.057 & 0.055 & 0.067 & 0.060 & \textbf{0.048} \\
BDD100K & CityScapes &Knet-DeepLab \cite{zhang2021knet} & 61.27 & 0.073 & 0.070 & 0.069 & 0.071  & 0.063 & 0.073 & 0.071 & \textbf{0.056}\\
BDD100K & CityScapes &Knet-SWIN-L \cite{zhang2021knet} & 67.46 & 0.080 & 0.078 & 0.077 & 0.077 & 0.071 & 0.079& 0.079 & \textbf{0.063}\\
BDD100K & CityScapes &ConvNeXt-L \cite{liu2022convnet} & 67.18 & 0.081 & 0.079 & 0.079 & 0.077 & 0.070 & 0.080 & 0.079 &  \textbf{0.065} \\
\hline


DAVIS-train & DAVIS-test & SegFormer-B4 \cite{xie2021segformer} &  89.33 & 0.033 & 0.031 & 0.031 & 0.031 & 0.055 & 0.098 & 0.032 & \textbf{0.024} \\
DAVIS-train & DAVIS-test & Segmenter-B \cite{strudel2021segmenter}&   81.35 & 0.080 & 0.077 & 0.076 &  0.075 & 0.103 & 0.121 & 0.078 & \textbf{0.049}\\
DAVIS-train & DAVIS-test & Knet-DeepLab \cite{zhang2021knet} &  83.12 & 0.076 & 0.072 & 0.073 & 0.073 & 0.098 & 0.107 & 0.075 &\textbf{0.047}\\
DAVIS-train & DAVIS-test & Knet-SWIN-B \cite{zhang2021knet}&   89.22 & 0.045 & 0.041 & 0.041 & 0.040 & 0.065& 0.101 & 0.044 & \textbf{0.029}\\
DAVIS-train & DAVIS-test & ConvNeXt-B \cite{liu2022convnet}& 89.00 & 0.044 & 0.040 & 0.041 & 0.042 & 0.067 & 0.100 & 0.044 & \textbf{0.031}\\
 \hline
 

SN-7-SP-train & SN-7-SP-test & SegFormer-B4 \cite{xie2021segformer} &  57.23 & 0.655 & 0.636 & 0.625 & 0.634 & 0.711 & 0.701 & 0.652 & \textbf{0.532} \\
SN-7-SP-train & SN-7-SP-test &Segmenter-B \cite{strudel2021segmenter}&  51.98&  0.766 & 0.751 & 0.748 & 0.740 & 0.796 & 0.789 & 0.763 & \textbf{0.687}\\
SN-7-SP-train & SN-7-SP-test &Knet-DeepLab \cite{zhang2021knet} & 55.58 & 0.704 & 0.688 & 0.679 & 0.681 & 0.768 & 0.720 & 0.696 &  \textbf{0.655}\\
SN-7-SP-train & SN-7-SP-test &Knet-SWIN-B \cite{zhang2021knet}& 59.42 & 0.609 & 0.576 & 0.559  & 0.563  & 0.631 & 0.618 & 0.605 & \textbf{0.504}\\
SN-7-SP-train & SN-7-SP-test &ConvNeXt-B \cite{liu2022convnet}& 58.69 & 0.611 & 0.570 & 0.564 & 0.566 &  0.629 & 0.633 &  0.608 &\textbf{0.517}\\
 \hline

SN-7-TS-train & SN-7-TS-test & SegFormer-B4 \cite{xie2021segformer} &  54.39 & 0.700 & 0.673 & 0.669 & 0.673 & 0.712 & 0.723 & 0.695 & \textbf{0.582} \\
SN-7-TS-train & SN-7-TS-test &Segmenter-B \cite{strudel2021segmenter}& 50.13 & 0.769 & 0.745 & 0.740 & 0.736 & 0.785& 0.777 & 0.755 & \textbf{0.624} \\
SN-7-TS-train & SN-7-TS-test &Knet-DeepLab \cite{zhang2021knet} & 56.73 & 0.651 & 0.622 & 0.621 & 0.619 & 0.689 & 0.675 & 0.647 & \textbf{0.576}\\
SN-7-TS-train & SN-7-TS-test &Knet-SWIN-B \cite{zhang2021knet}& 62.42 & 0.619 & 0.587 & 0.579 & 0.584 & 0.630 & 0.621 & 0.609 & \textbf{0.551} \\
SN-7-TS-train & SN-7-TS-test &ConvNeXt-B \cite{liu2022convnet}& 62.14 & 0.623 & 0.578 & 0.571 & 0.573 & 0.647 & 0.638 & 0.620 & \textbf{0.549} \\
 \hline
BraTS-train & BraTS-test & SegFormer-B4 \cite{xie2021segformer} & 45.81 & 0.155 & 0.137 & 0.135 & 0.135 &0.129 &0.150 & 0.151 & \textbf{0.120}\\
BraTS-train & BraTS-test &Segmenter-B \cite{strudel2021segmenter}& 44.57 & 0.154 & 0.149 & 0.146 & 0.147 & 0.135 & 0.142 & 0.150 & \textbf{0.124}  \\
BraTS-train & BraTS-test &Knet-DeepLab \cite{zhang2021knet} &  46.05 & 0.188 &0.179 & 0.176 & 0.176 & 0.158 & 0.187 & 0.185 & \textbf{0.148}\\
BraTS-train & BraTS-test &Knet-SWIN-B \cite{zhang2021knet}&  47.99 & 0.155 & 0.146 & 0.143 & 0.144& 0.132 & 0.157 & 0.152 & \textbf{0.123} \\
BraTS-train & BraTS-test &ConvNeXt-B \cite{liu2022convnet}&  48.40 & 0.169 & 0.159 & 0.154 & 0.155& 0.136 & 0.168 & 0.165 & \textbf{0.121} \\
 \hline
SYNTHIA & CityScapes & SegFormer-B5 \cite{xie2021segformer}& 33.12 & 0.286 & 0.264 & 0.262 & 0.263 & 0.230 & 0.288 & 0.284 & \textbf{0.218}\\
SYNTHIA & CityScapes &Segmenter-L \cite{strudel2021segmenter}&  32.33 & 0.254 & 0.235 & 0.232 & 0.230 & 0.217 & 0.280 & 0.252 & \textbf{0.203}\\
SYNTHIA & CityScapes &Knet-DeepLab \cite{zhang2021knet} &  30.19 & 0.311 & 0.298 & 0.296 & 0.296 & 0.272 & 0.305 & 0.309 & \textbf{0.241} \\
SYNTHIA & CityScapes &Knet-SWIN-L \cite{zhang2021knet} &  34.31 & 0.281 & 0.273 & 0.270 & 0.270 & 0.239 & 0.288 &  0.279 & \textbf{0.226}\\
SYNTHIA & CityScapes &ConvNeXt-L \cite{liu2022convnet} & 34.79 & 0.287 & 0.267 & 0.263 & 0.264 & 0.230 & 0.299 & 0.284  &\textbf{0.211}\\


 \hline
\end{tabular}
\vspace{-1em}
\end{table*}

\vspace{-1em}

\subsubsection{Ablation study}
  \vspace{-0.5em}

To analyze the effect of selector on selective scaling, we conduct ablation study by varying misprediction detection accuracy, integrating with different existing calibrators, and examining calibration error across regions. 

\begin{figure}
    \centering
    \includegraphics[width=0.18\textwidth]{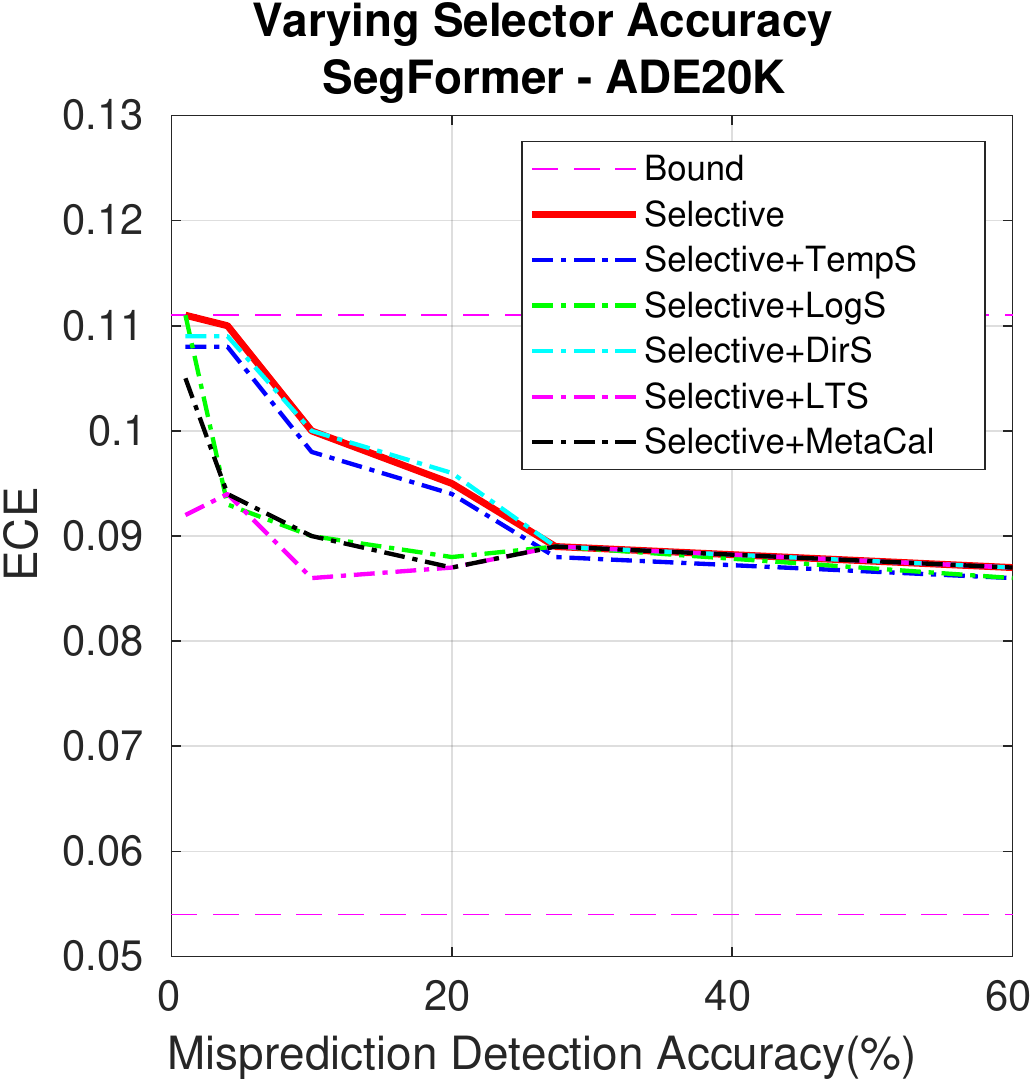} \hspace{4mm}
    \includegraphics[width=0.18\textwidth]{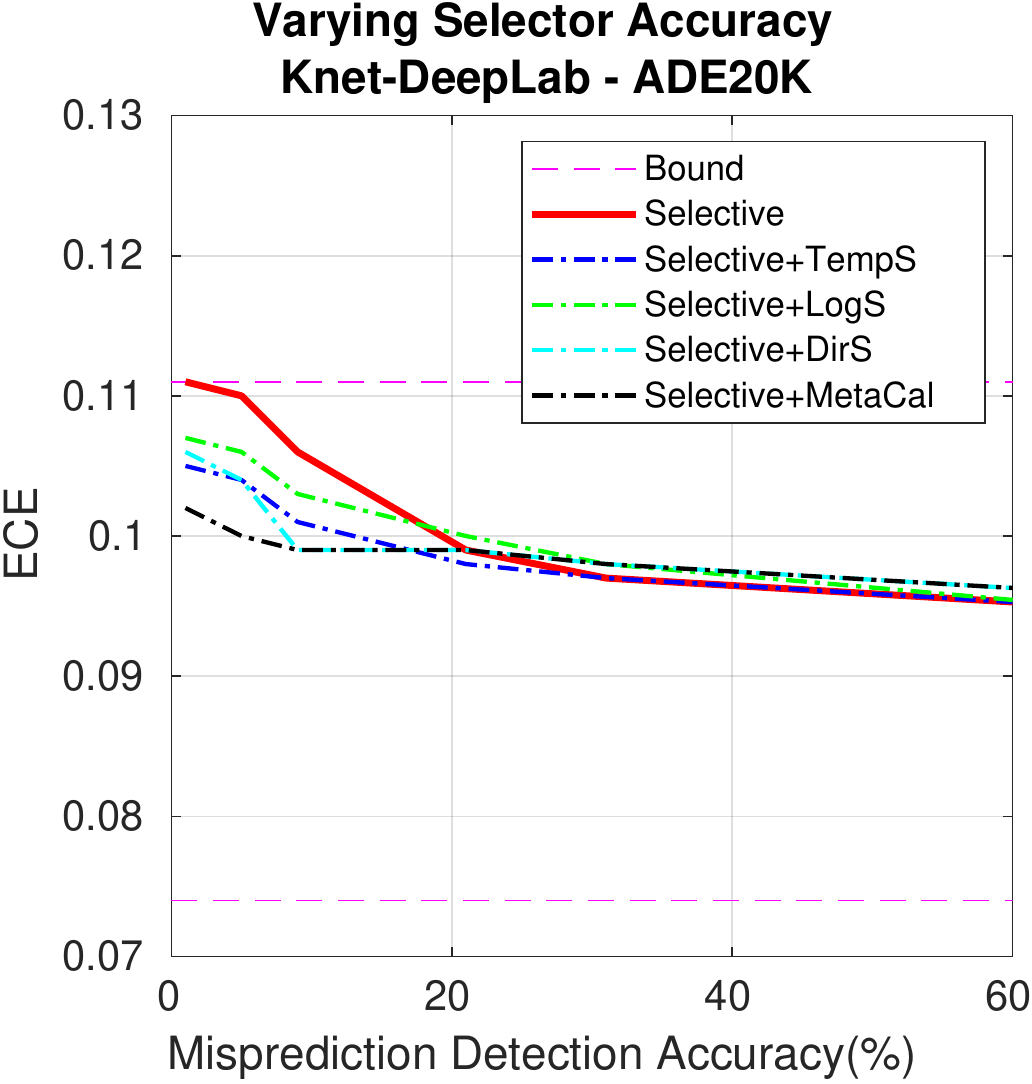}
    \caption{The effect of selector accuracy on calibration. Misprediction detection accuracy denotes the percentage of misprediction detected by selector. The lower bound (optimal case) is 100$\%$ misprediction detected for scaling. The upper bound (worst case) is 0$\%$, \ie, none misprediction detected is used for calibration.}
    \vspace{-1em}
    \label{fig:acc_selector}
\end{figure}

\textbf{Selective scaling considering selection accuracy.} Figure~\ref{fig:acc_selector} shows that with the increase in selection accuracy, ECE decreases. This implies that when more mispredictions are detected and scaled, model calibration becomes better. It justifies the effectiveness of selective scaling.

\textbf{Selective scaling with existing calibrators.} We also find that selective scaling can boost existing calibrators, vice versa. Figure~\ref{fig:acc_selector} shows that the integration with selective scaling helps existing calibrators to improve model calibration. When detection accuracy is lower, like $<20\%$, existing calibrators can help improve selective scaling.



\begin{figure}[h]
    \centering
    \includegraphics[width=0.40\textwidth]{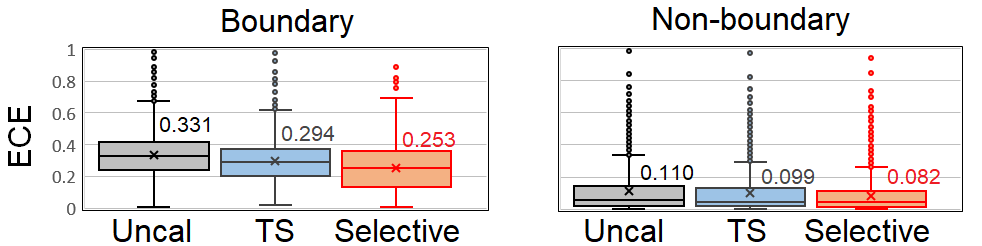}
     \vspace{-0.5em}
   \caption{SegFormer-B5 ECE comparison in ADE20K between boundary and non-boundary pixels among uncalibration, temperature scaling, and selective scaling. Crosses and dots show means and outliers. The numbers are the means.}
    \vspace{-1em}
    \label{fig:ob}
\end{figure}

\textbf{Selective scaling across regions.} Figure \ref{fig:ob} shows segmentation calibration error statistics across regions. Top and bottom bars denote maximum and minimum ECE while top, central, and bottom lines of the box indicate $25\%$, $50\%$, and $75\%$ of ECE distribution. It shows that: 1) boundary pixels are less calibrated than non-boundary pixels within an object bounding box; 2) selective scaling consistently yields lower mean ECE across both types of pixels. 

\subsection{Limitation}
Logit smoothing can cause entropy increase because post-hoc over-confidence calibration, such as temperature scaling, is equivalent to entropy maximization under certain logit constraints \cite{guo2017calibration}. For example, original Meta-cal leads to significant entropy increase due to randomizing certain prediction outputs. Our method alleviates this issue, but still yields higher entropy. To resolve it, careful temperature tuning is a trade-off scheme for entropy improvement.

  \vspace{-0.5em}
\section{Conclusion}

We analyzed semantic segmentation model calibration and proposed a simple but effective algorithm of selective scaling. We revealed the possible factors of segmentation model miscalibration. Given the analysis, we designed selective scaling to scale logits separately, and more focus on mispredicted logit smoothing. We performed extensive experiments on state-of-the-art models across various benchmarks. Both in-domain and domain-shift experiments justify the effectiveness of selective scaling. We also presented our observations for segmentation calibration research.

{\small
\bibliographystyle{ieee_fullname}
\bibliography{egbib}
}

\end{document}


\title{Supplementary \\ On Calibrating Semantic Segmentation Models: Analyses and An Algorithm}

\maketitle

\section{Uncertainty of Semantic Segmentation}

\begin{figure}[h]
    \centering
    \subcaptionbox{\label{sfig:aa}}{\includegraphics[width=0.22\textwidth]{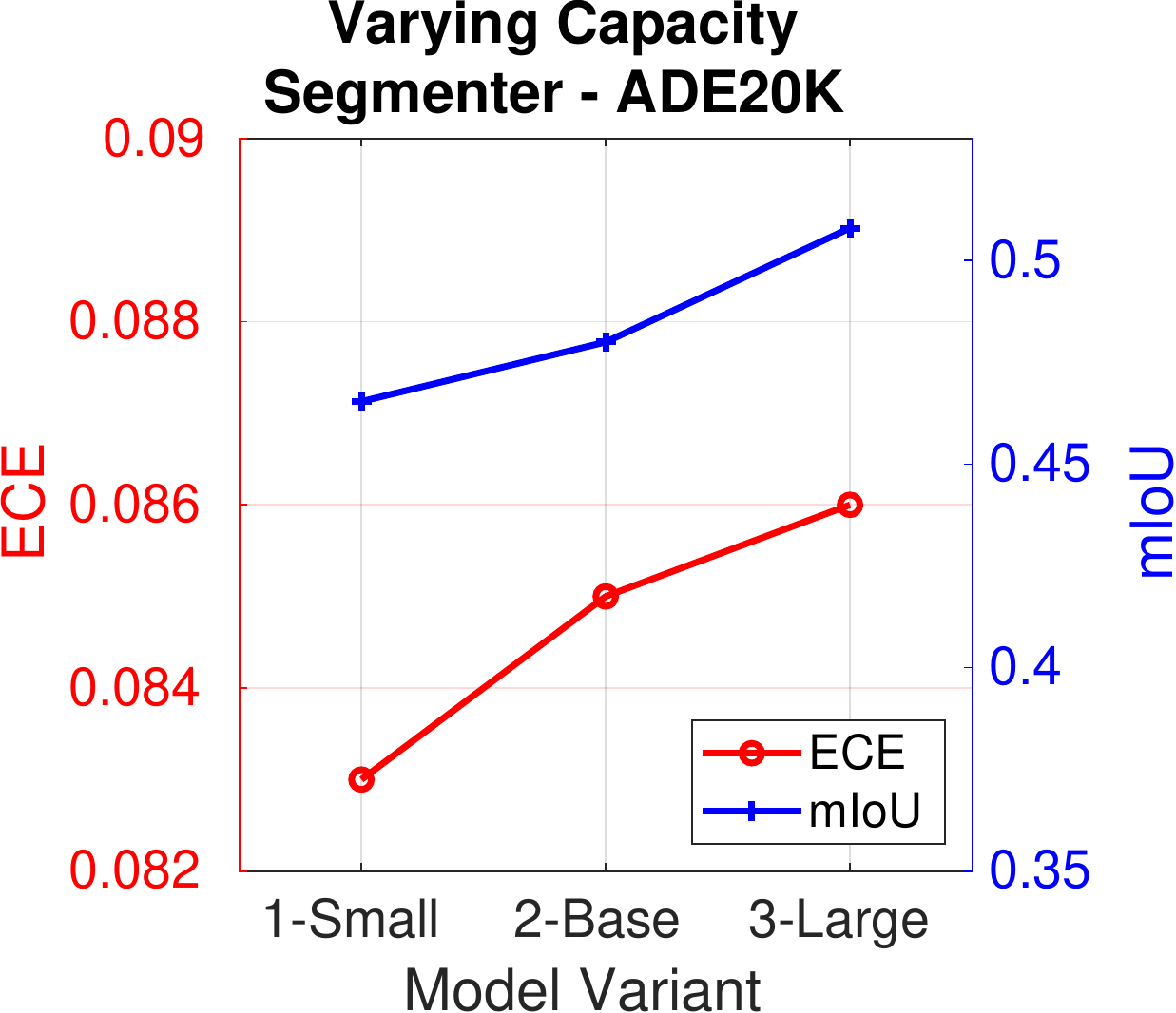}}
    \subcaptionbox{\label{sfig:ab}}{\includegraphics[width=0.22\textwidth]{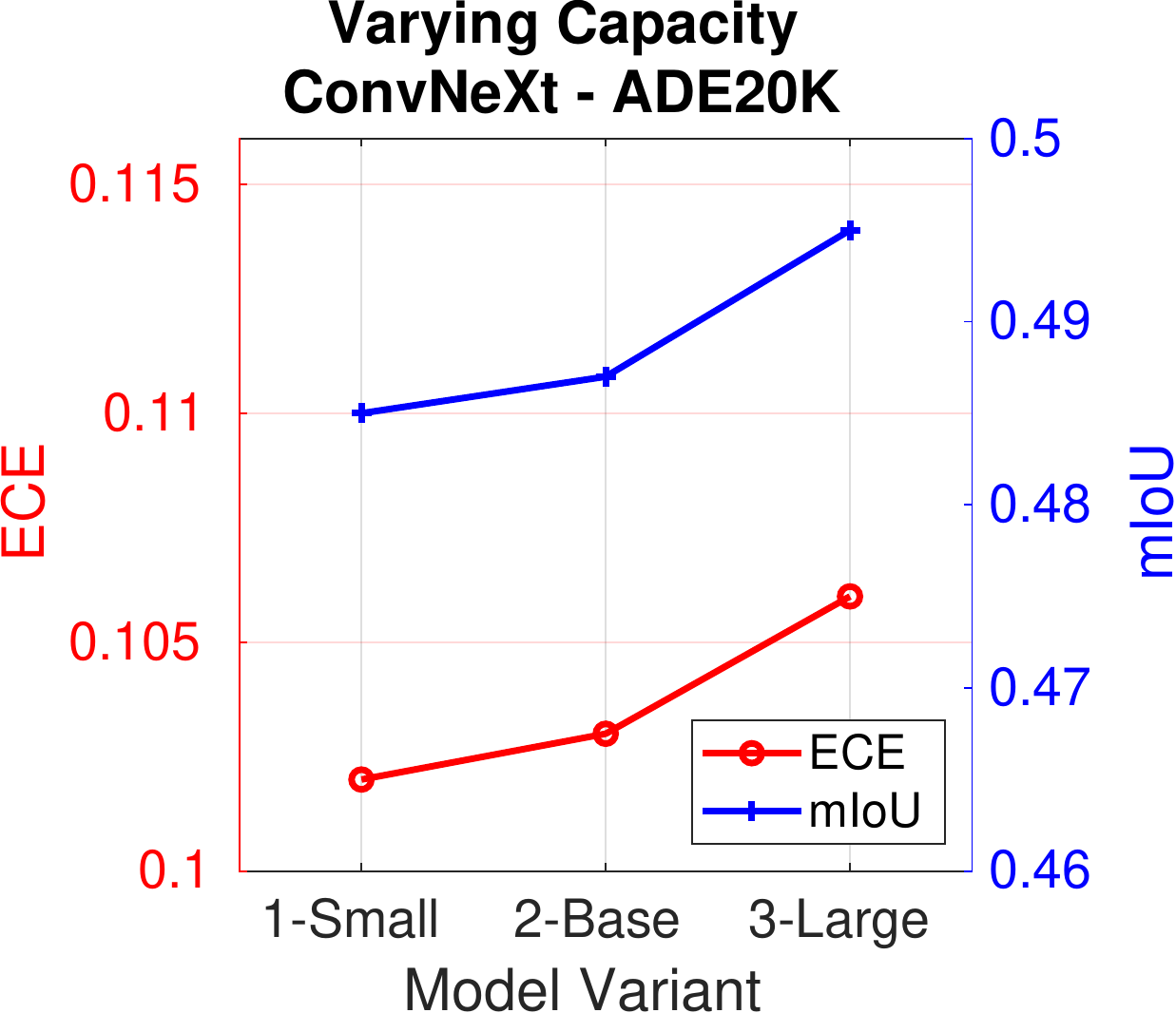}}
    \caption{The effect of model capacity. Image-based ECE is employed to report model miscalibration. Model calibration error (ECE) tend to increase as model size increases given the observations from Segmenter and ConvNeXt. }
    \label{fig:model_capacity}
\end{figure}

\begin{figure}[h]
    \centering
    \subcaptionbox{\label{sfig:ba}}{\includegraphics[width=0.22\textwidth]{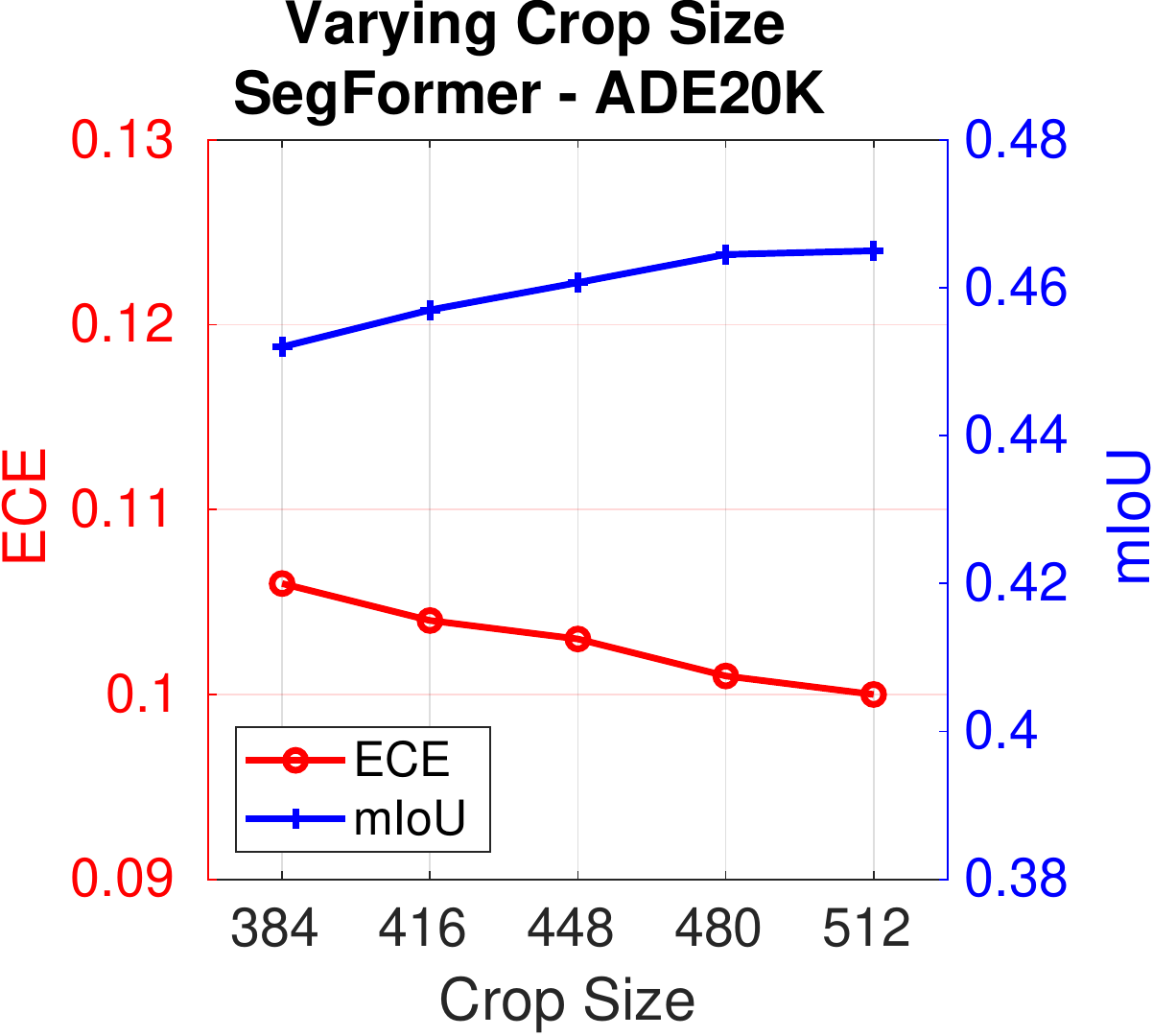}}    
    \subcaptionbox{\label{sfig:bb}}{\includegraphics[width=0.22\textwidth]{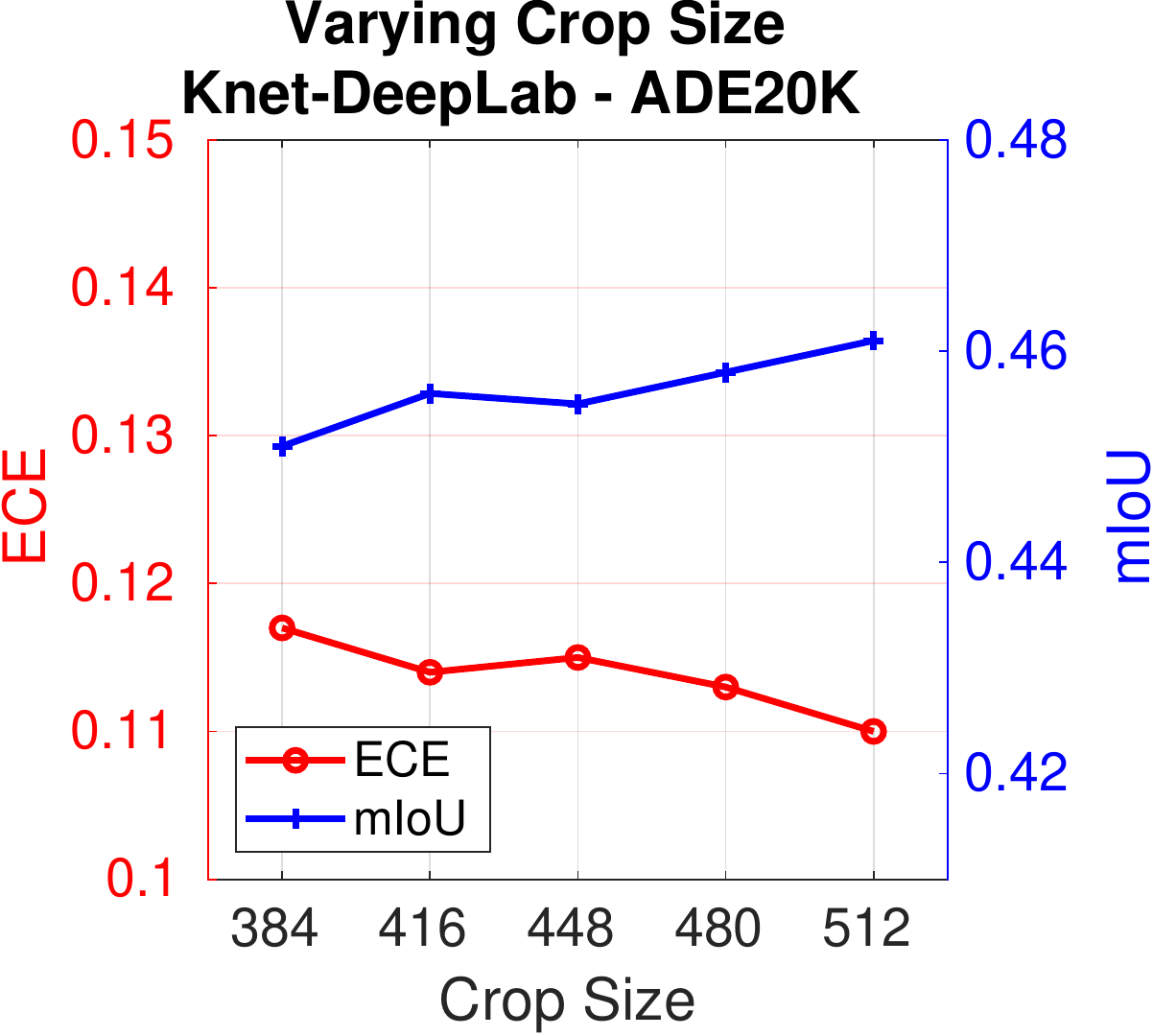}}
    \subcaptionbox{\label{sfig:bc}}{\includegraphics[width=0.22\textwidth]{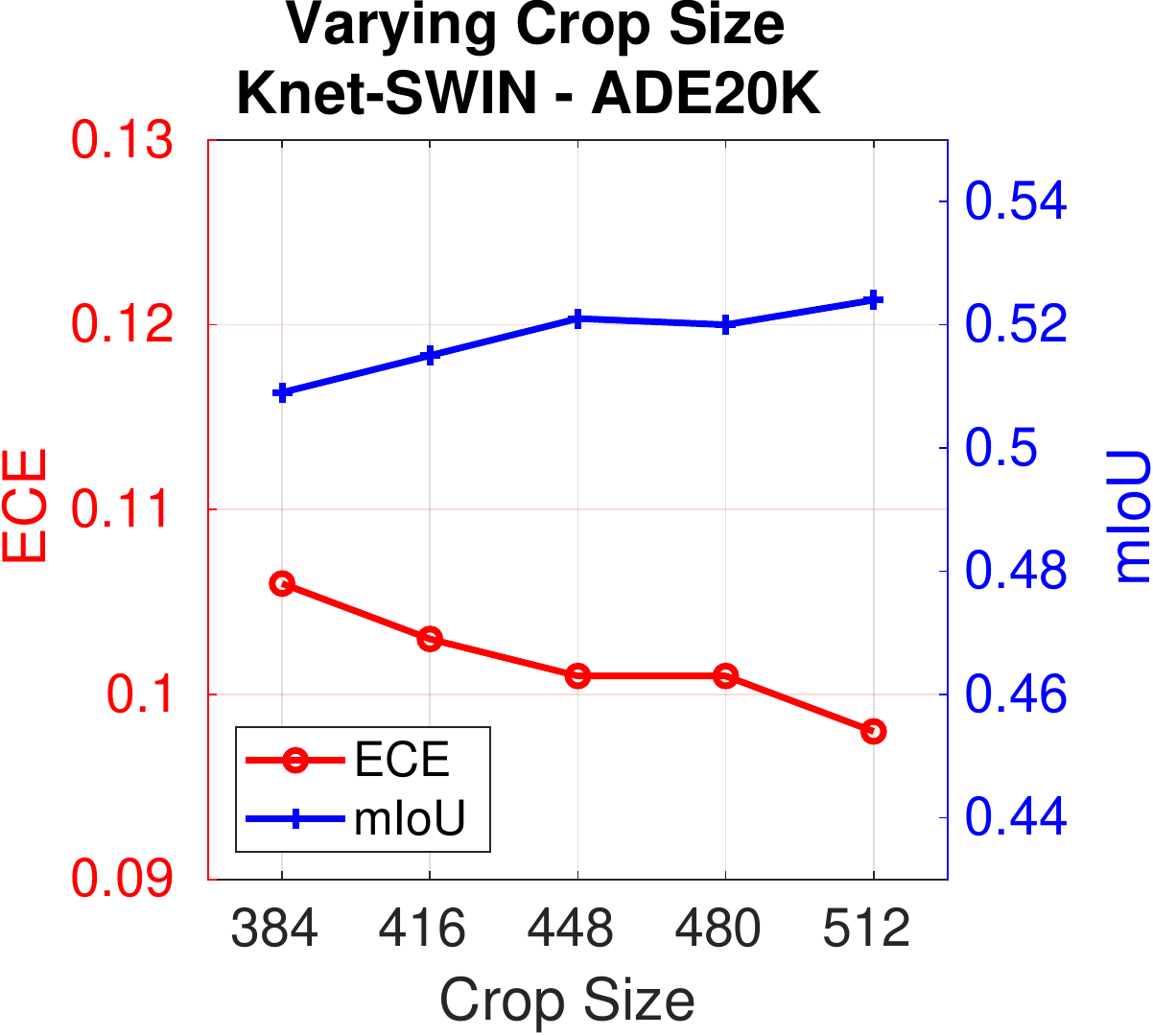}}
    \subcaptionbox{\label{sfig:bd}}{\includegraphics[width=0.22\textwidth]{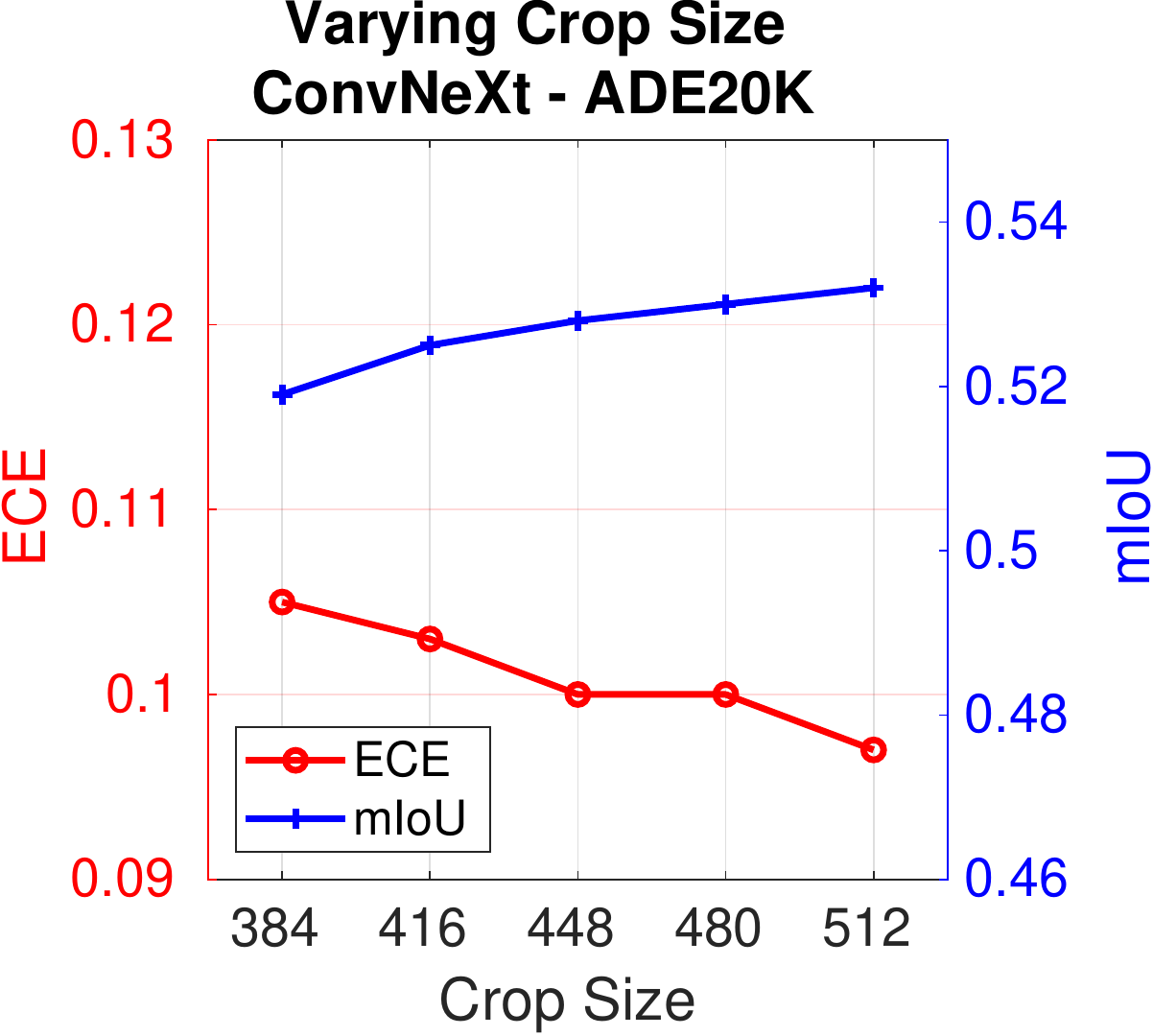}}
    \caption{The effect of crop size. Image-based ECE is used to report miscalibration. Miscalibration tends to increase as crop size increases given the observations across four models. }
    \label{fig:model_cropsize}
\end{figure}

\vspace{20in}

\begin{figure}[h]
    \centering
    \subcaptionbox{\label{sfig:ca}}{\includegraphics[width=0.22\textwidth]{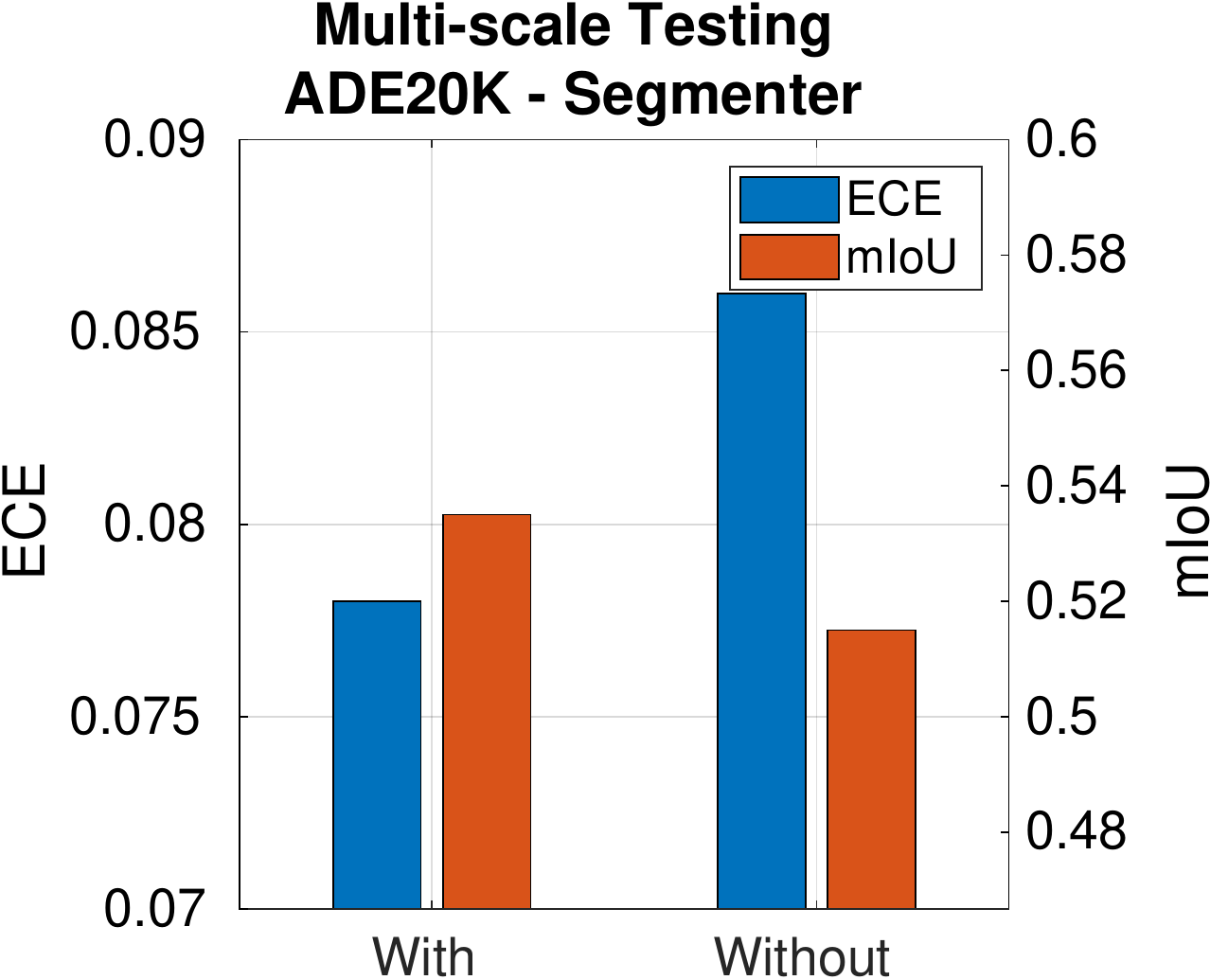}} 
    \subcaptionbox{\label{sfig:cb}}{\includegraphics[width=0.22\textwidth]{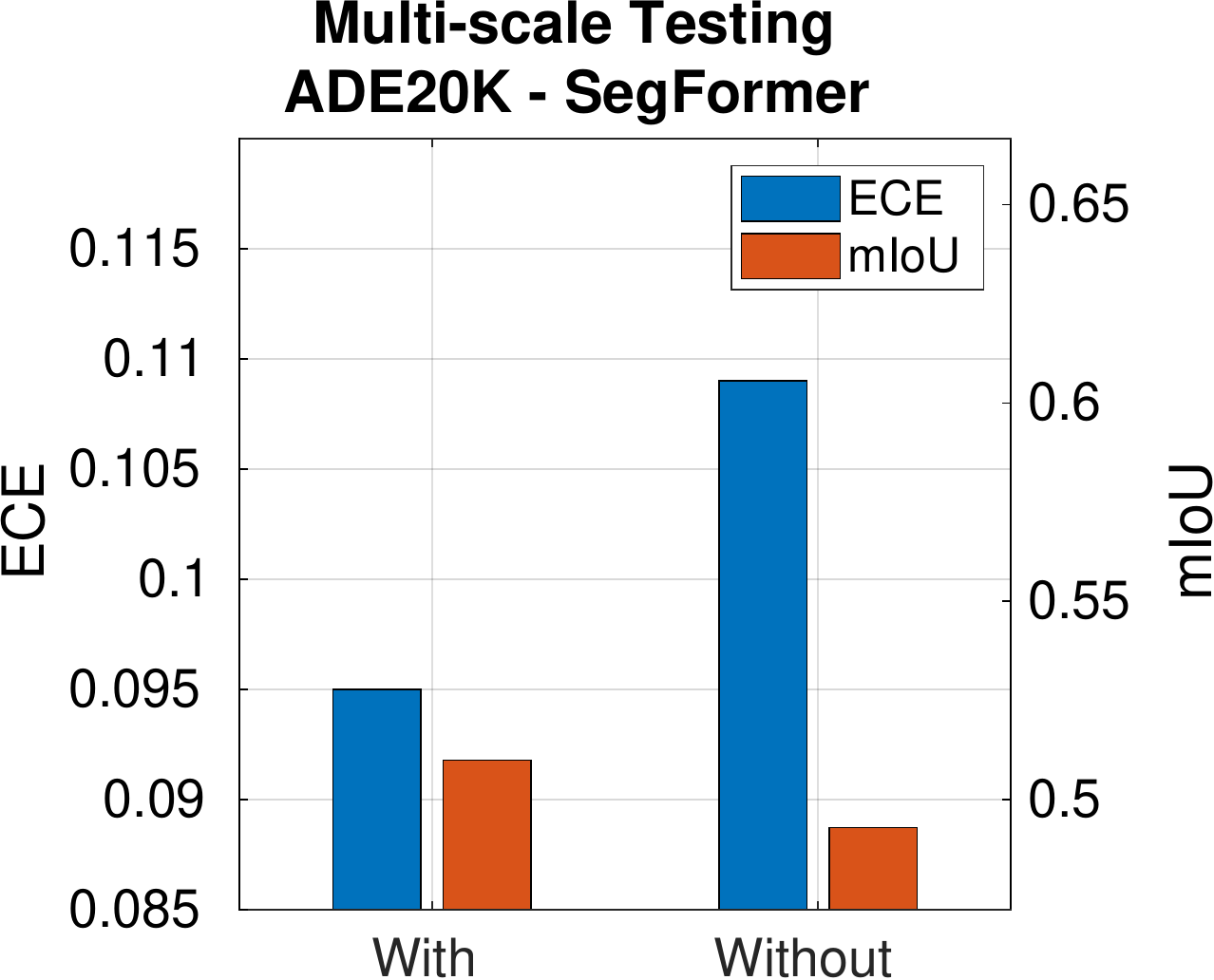}}    
    \subcaptionbox{\label{sfig:cc}}{\includegraphics[width=0.22\textwidth]{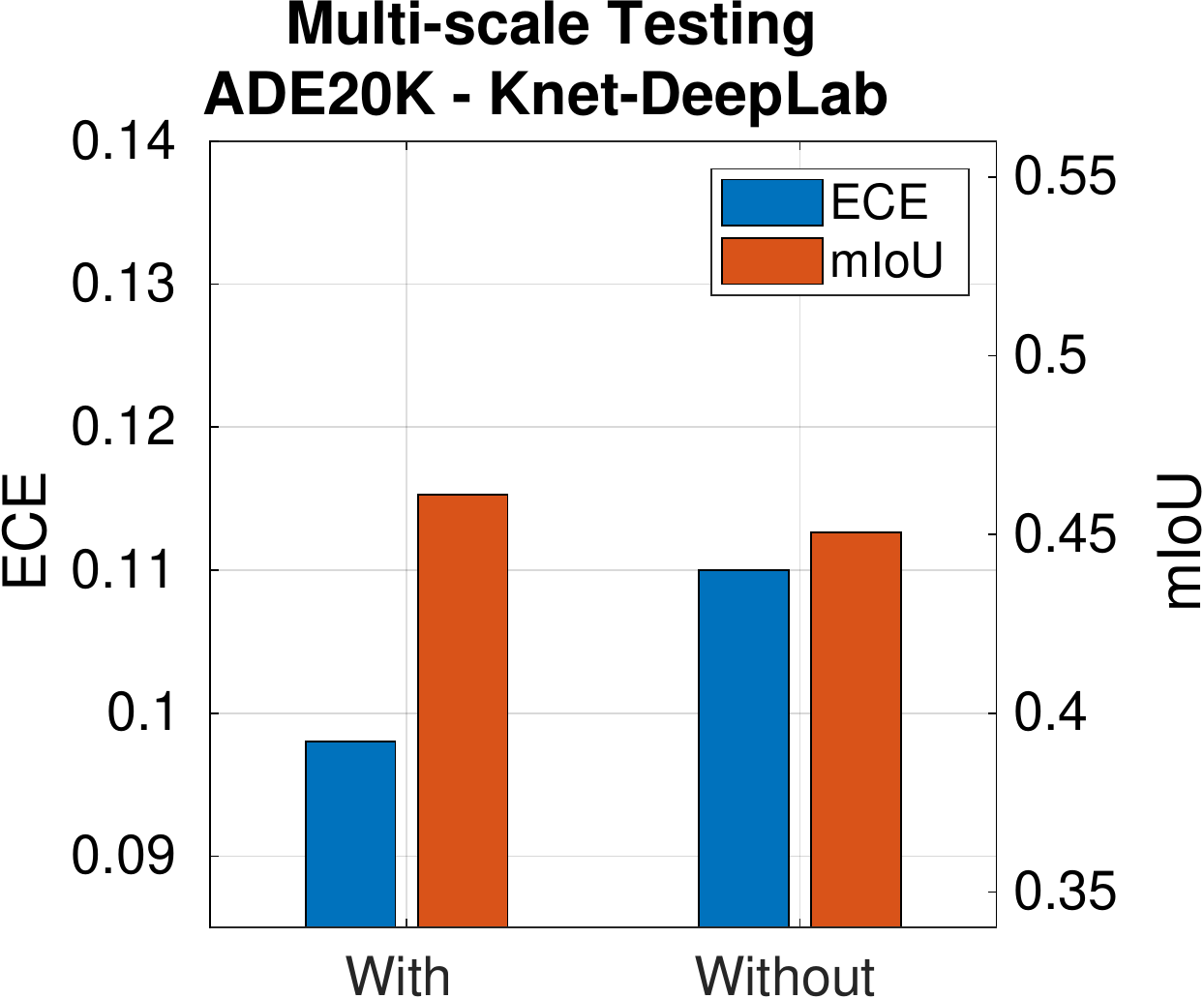}}
    \subcaptionbox{\label{sfig:cd}}{\includegraphics[width=0.22\textwidth]{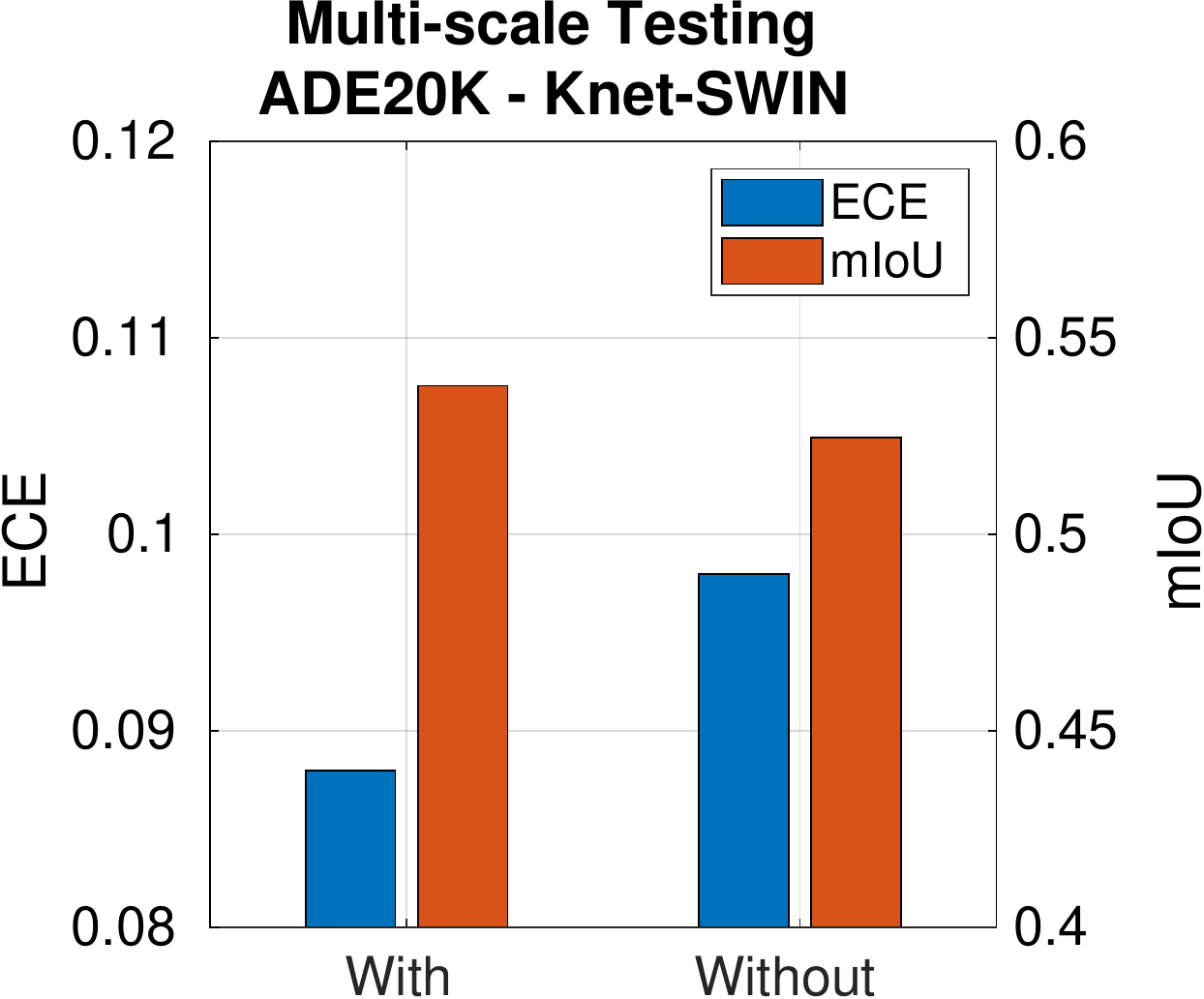}}
    \caption{The effect of multi-scale testing. Image-based ECE is adopted to compare different testing strategies. From the observations, all four models show that mIoU increases, but ECE decreases when multi-testing is employed. }
    \label{fig:model_cropsize}
\end{figure}

\begin{figure}[h]
    \centering
    \includegraphics[width=0.48\textwidth]{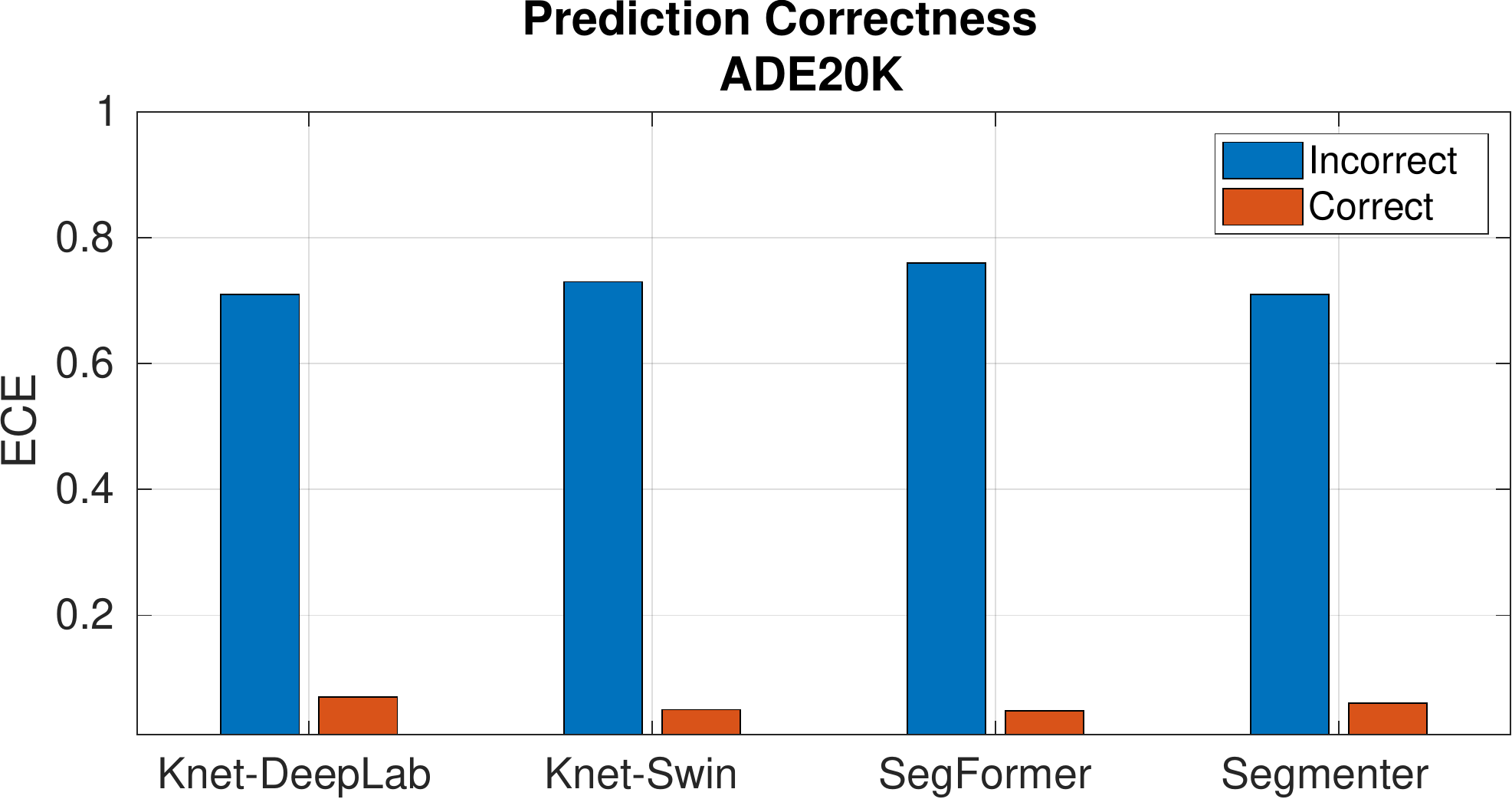}   
    \caption{The effect of prediction correctness. Image-based ECE is adopted to assess miscalibration. Misprediction contributes more to miscalibration given the ECEs across four models. }
    \label{fig:model_cropsize}
\end{figure}

\vspace{20in}

\newpage

\section{Reliability Diagrams}

\begin{figure}[h]
    \centering
    \includegraphics[width=0.48\textwidth]{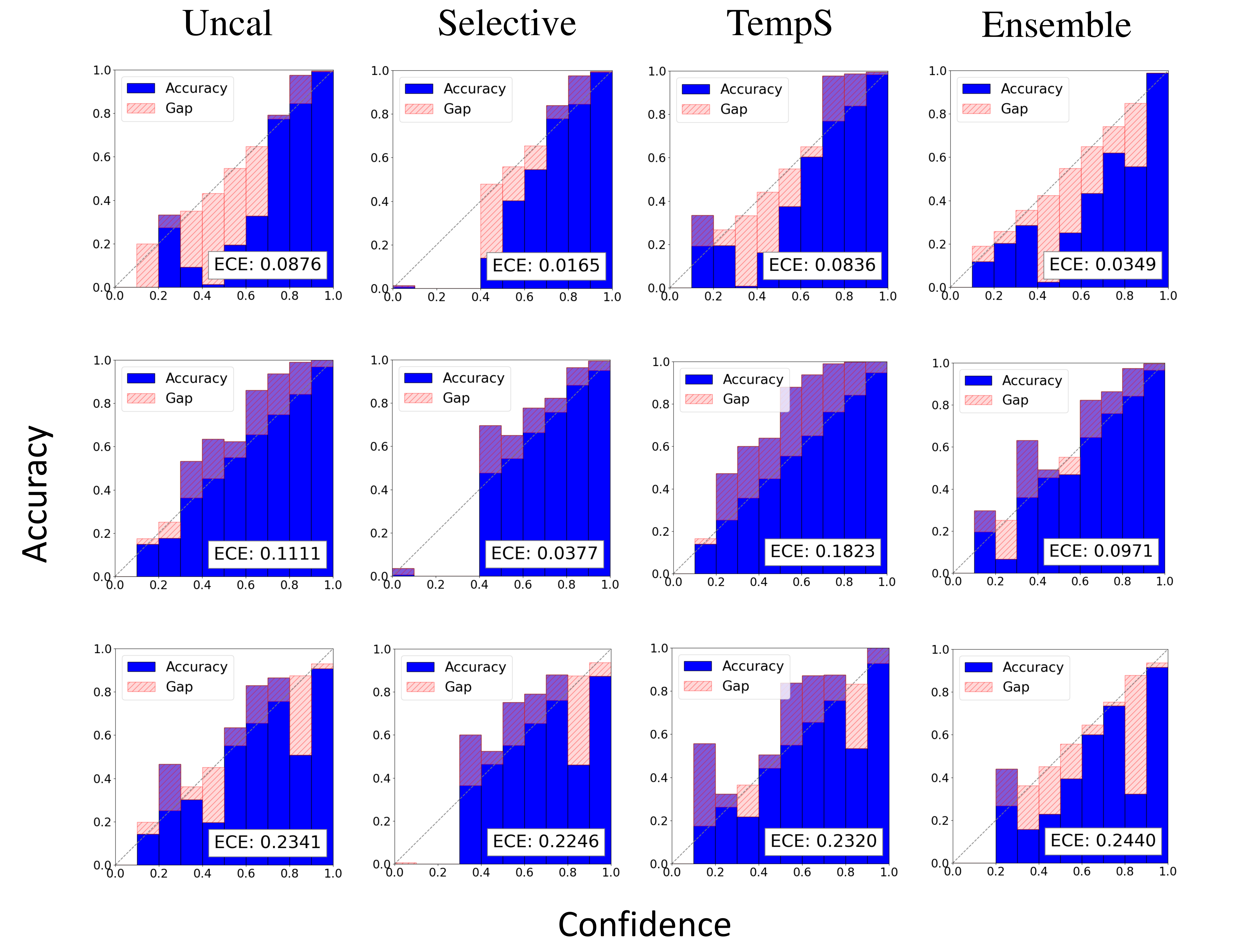}   
    \caption{Reliability diagrams of visualized ECE for Segmenter-L on COCO-164K. Different calibration methods are compared across three randomly selected individual images. Selective scaling consistently outperforms temperature scaling and ensembling. }
    \label{fig:reliability_coco}
\end{figure}

\section{Spatial Statistics on Calibration Errors}

 \begin{figure}[h]
    \centering
    \includegraphics[width=0.45\textwidth]{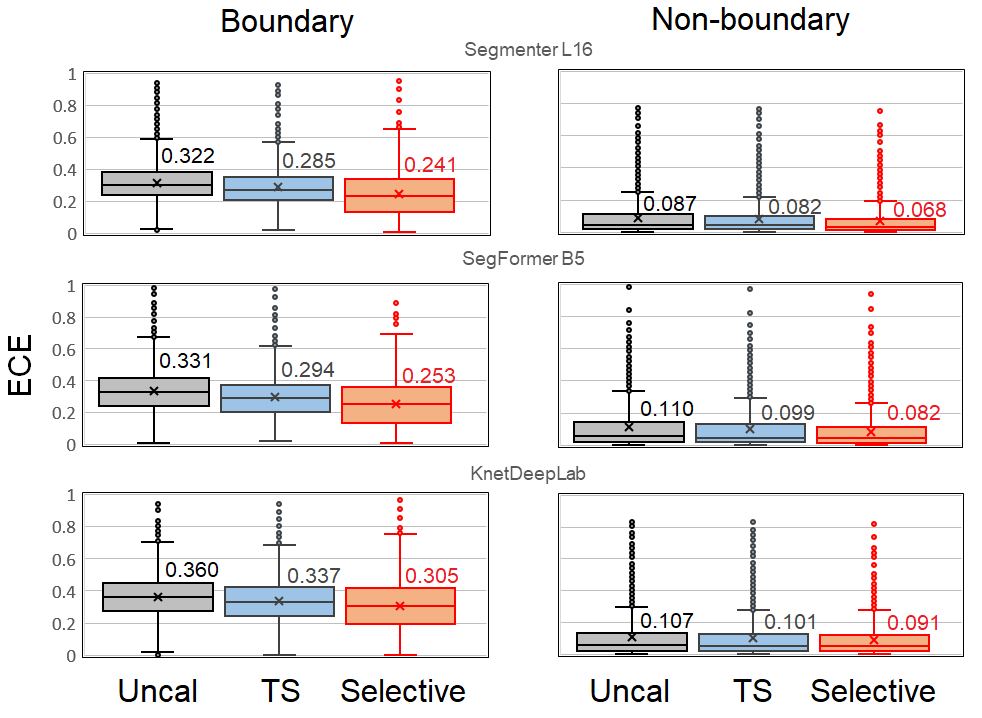}
     \vspace{-1em}
   \caption{Calibration error (ECE) comparison in ADE20K between boundary and non-boundary pixels with boxplot of image-wise ECEs. Top and bottom bars denote maximum and minimum ECE while top, central, and bottom lines of the box indicate $25\%$, $50\%$, and $75\%$ of ECE distribution. Crosses and dots show means and outliers. The numbers are the means. We conduct a comparative study on calibration error among uncalibration, temperature scaling, and selective scaling.}
    \vspace{-0.75em}
    \label{fig:ob}
\end{figure}

\vspace{5in}

\section{Selective Scaling Calibrator Training Setting}

\begin{table}[h]
\small
\setlength\tabcolsep{2pt}
\centering
\caption{Hyperparameter information.}
\vspace{-1em}
\begin{tabular}{c c | c c | c c }
\hline
BatchSize  & 20 & Epoch & 40 & LR & 0.001 \\ 
Optimizer & AdamW & Decay & 1e-6 & Loss & CrossEntropy\\
\hline
Hyperparameter & \multicolumn{5}{c}{1e10 when ACC$_{misprediction}>50\%$} \\
T2 & \multicolumn{5}{c}{2 when ACC$_{misprediction}<35\%$ } \\ \hline
\end{tabular}
 \label{tab:hyper}
\end{table}
